\documentclass{article}

\usepackage{multirow}

    \usepackage[nonatbib, preprint]{neurips_2023}

\usepackage{amsmath}
\usepackage[utf8]{inputenc} % allow utf-8 input
\usepackage[T1]{fontenc}    % use 8-bit T1 fonts
\usepackage{hyperref}       % hyperlinks
\usepackage{url}            % simple URL typesetting
\usepackage{booktabs}       % professional-quality tables
\usepackage{amsfonts}       % blackboard math symbols
\usepackage{nicefrac}       % compact symbols for 1/2, etc.
\usepackage{microtype}      % microtypography
\usepackage{xcolor}         % colors
\usepackage{graphicx}
\usepackage{colortbl}
\usepackage{subcaption}

\title{Towards a Robust Framework for NeRF Evaluation}

\author{%
  Adrian Azzarelli \\
 Visual Information Laboratory\thanks{This work was funded by the UKRI MyWorld Strength in Places Programme (SIPF00006/1).}\\
 University of Bristol\\
  \texttt{a.azzarelli@bristol.ac.uk} \\
  \And
  Nantheera Anantrasirichai \\
Visual Information Laboratory\\
  University of Bristol \\
  \texttt{n.anantrasirichai@bristol.ac.uk} \\
  \And
  David R Bull \\
Visual Information Laboratory\\
  University of Bristol \\
  \texttt{dave.bull@bristol.ac.uk} \\
}

\begin{document}

\maketitle

\begin{abstract}
    Neural Radiance Field (NeRF) research has attracted significant attention recently, with 3D modelling, virtual/augmented reality, and visual effects driving its application. While current NeRF implementations can produce high quality visual results, there is a conspicuous lack of reliable methods for evaluating them. Conventional image quality assessment methods and analytical metrics (e.g. PSNR, SSIM, LPIPS etc.) only provide approximate indicators of performance since they generalise the ability of the entire NeRF pipeline. Hence, in this paper, we propose a new test framework which isolates the neural rendering network from the NeRF pipeline and then performs a parametric evaluation by training and evaluating the NeRF on an explicit radiance field representation. We also introduce a configurable approach for generating representations specifically for evaluation purposes. This employs ray-casting to transform mesh models into explicit NeRF samples, as well as to ``shade" these representations. Combining these two approaches, we demonstrate how different ``tasks" (scenes with different visual effects or learning strategies) and types of networks (NeRFs and depth-wise implicit neural representations (INRs)) can be evaluated within this framework. Additionally, we propose a novel metric to measure task complexity of the framework which accounts for the visual parameters and the distribution of the spatial data. Our approach offers the potential to create a  comparative objective evaluation framework for NeRF methods.
\end{abstract}

\section{Introduction}\label{sec:intro}
Neural Radiance Fields (NeRFs) are a class of neural network capable of learning a 3-D scene from a reasonably small number of images captured from different viewpoints \cite{mildenhall2021nerf}. They perform neural rendering with a focus on \textit{view-dependant novel view synthesis} and have overcome a number of significant challenges associated with automated 3-D capture \cite{gao2022nerf, xie2022neural}.  They provide automated rendering without the need for compiling \textit{shaders}\footnote{Shaders are functions for simulating different physical-visual spaces, like \textit{light fields}.} and, in conjunction with  image-pose estimation tools and methods for sampling points in space,  are able to generate realistic 3-D representations of a target scene \cite{tancik2023nerfstudio,levy2023seathru, jambon2023nerfshop}. 

Despite their popularity, benchmarking the performance of NeRFs remains problematic - especially as state-of-the-art methods become closer in performance, \cite{xie2022neural}. Most contemporary comparisons employ conventional image quality assessment metrics alongside subjective results based on selected ``novel view'' images.  Crucially, a NeRF is a form of Implicit Neural Representation (INR) that models two visual features: volumetric colour and density. Other INRs model visual and spatial (geometric) features \cite{kong2023vmap,chng2022garf,yu2021plenoxels}, other than colour and density. These geometric features can be exploited in the NeRF pipeline \cite{kurz2022adanerf,neff2021donerf} as a basis for sampling volumetric surfaces for rendering. Thus by involving the entire pipeline in quality assessment, information may be lost or masked due to the accumulation of errors from different pipeline components. This was discussed in \cite{wang2023benchmarking}, where the issue of corrupted images within the pipeline was addressed leading to more stable benchmarks. However, as image-based metrics only evaluate the prediction quality via a 2-D projection, this can still result in loss of information about the accuracy of spatial samples in relation to their distribution in volumetric space. Comparing the performance of different NeRF methods can thus be challenging when the process of sampling NeRF inputs is specific to the rendering network. This is evidenced in Nerfacto \cite{tancik2023nerfstudio} and Instant-NGP (INGP)  \cite{muller2022instant} where sampling approaches are adapted for higher learning rates (i.e. fast methods). 

\begin{figure}[t!]
    \centering
    \includegraphics[width=\textwidth]{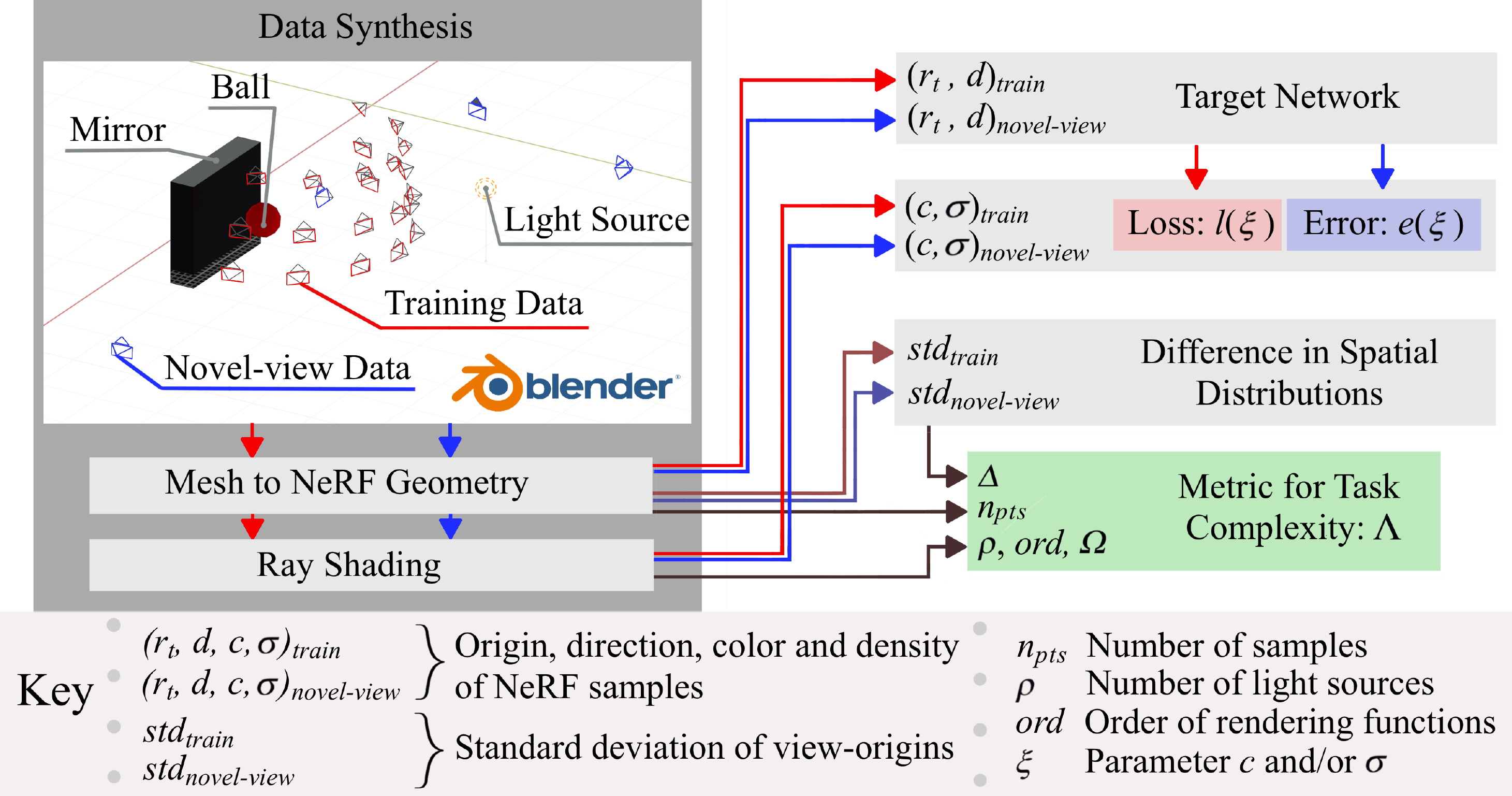}
    \caption{Our proposed framework for evaluation. The \textit{Target Network} is the rendering network generating NeRF colour and density fields. The data synthesis employs the \textit{Mesh to NeRF} and \textit{Ray Shading} modules to input the target network and to evaluate the colour and density predictions, respectively. For other INRs, the inputs and outputs will be different so may require some modification.}
    \label{fig:framework}
\end{figure}

In this paper, we proposed a novel framework that provides a basis for robust and consistent objective parametric evaluation, addressing the difficulty of approximating a 3-D scene using NeRFs. We validated this framework by evaluating the performance of different networks on the same scene with various material effects. The main contributions of our work are as follows:
\begin{itemize}
    \item  We proposed a new metric, the \textit{Whole-scene Average Prediction Error} (WAPE), to evaluate the performance of a INR rendering network. It calculates the mean absolute error of the learnable outputs, e.g. colour and density for the NeRF methods, against the ground truth.
    \item We demonstrated that the ground truths can be accurately represented as a synthetic radiance field.  This is achieved by transforming mesh-based representations and applying ray tracing to enhance the quality of visual features.
    \item We also proposed a new metric for evaluating \textit{task complexity}. This takes into account the number of input samples, the relative distribution of novel views and training views, and the functional complexity of the chosen ray tracing algorithm(s).
    \item Our proposed framework, shown in Figure \ref{fig:framework}, combines these methods supporting adaptation for different types of evaluation. We validated this by evaluating the performance of an INR on with a different learning objective and neural representation to NeRF.
\end{itemize}

\begin{figure}[t]
    \centering
    \includegraphics[width=0.75\textwidth]{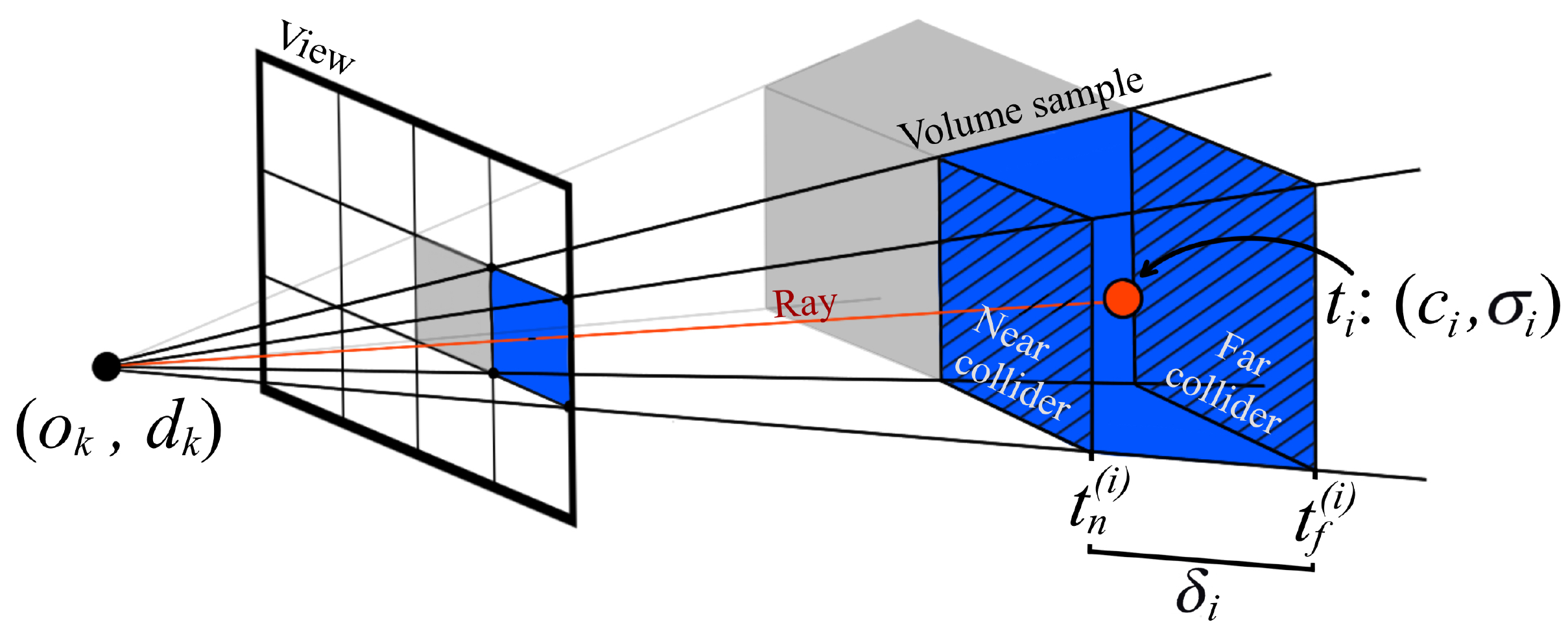}
    \caption{Example of a NeRF volumetric sample using Equation \ref{eq: NeRF colouring} to render volume samples falling along the same ray}
    \label{fig: Vol Rep}
\end{figure}

The remainder of this paper is organised as follows.  Section \ref{sec: problemstate} discusses existing work and highlights a number of paradigms for evaluation. Section \ref{sec:main} describes the proposed WAPE evaluation method and defines a metric for task complexity. Section \ref{sec:sythesizing} describes our methodologies for synthesising ground truth. Section \ref{sec: experiments} presents the experiments for evaluating the performance of both INRs and NeRFs. Section \ref{sec: limits} concludes the paper, discusses the limitations of our approach and potential for extension. We have provided code which works with NeRFStudio \cite{tancik2023nerfstudio}.

%%%%%%%%%%%%%%%%%%%%%%%%%%%%%%%%%%%%%%%%%%%%%%%%%%%%%%%%%%%%%%%%%%%%%%%%%%%%%%%%%%%%%%%%%%%%%%%%%%%%%%%%%%%%%%%%%%%%%%%%%%%%%

\section{Related Work and Problem Statement}\label{sec: problemstate}

\paragraph{Benchmarks and Current Objective Metrics} NeRF architectures have been extensively studied \cite{gao2022nerf, xie2022neural}\footnote{Xie et al. \cite{xie2022neural} provides a \textit{neural fields} search engine.} with the aim of improving training speed \cite{muller2022instant,yu2021plenoctrees}, image quality \cite{mildenhall2022nerf} and scene coverage \cite{zhang2020nerf++}. Less work has however been reported on  improving performance evaluation, with conventional image quality metrics typically employed  \cite{xie2022neural,gao2022nerf,tancik2023nerfstudio}. These are discussed  below. 

Peak Signal to Noise Ratio (PSNR) is a universal metric that provides a color-wise evaluation,  aligning well with the NeRF objective of learning spatially dependent color values. Structural Similarity Index Measure (SSIM) \cite{wang2004image}  measures the similarity between two images based on their structural information, luminance, and contrast. As SSIM computations are performed on image patches, they allow for some misalignment between the synthesized and reference images. This is helpful for evaluation as there may be variations between the NeRF camera model and the real camera used for capturing the training images. Finally, Learned Perceptual Image Patch Similarity (LPIPS) \cite{zhang2018unreasonable} has gained popularity in areas such as frame interpolation, measuring the similarity between features of two images extracted from a pretrained network. 

While widely used, these metrics have limitations when used to compare the performance of NeRF algorithms. This was demonstrated by Wang et al. \cite{wang2023benchmarking} who examined how image localisation/mapping and image-corruption from using real images influences image-based evaluation. They mitigated view misalignment and introduced extra metrics to measure the model's robustness in novel view synthesis tasks. De Luigi et al. \cite{de2023scannerf} also proposed benchmarking with \textit{scalable} scenes and provided a method for generating a \textit{Digital Twin} for evaluation.

% \paragraph{Pipeline Evaluation} 
Additional statistics can be used when comparing state-of-the-art methods to highlight the performance of various components involved in the NeRF pipeline. For example, the distribution of predicted samples along a ray \cite{lee2023dense, uy2023scade} can be helpful for indicating the quality of a sampling strategy, especially when specific sampling strategies are required for specific NeRF networks. However, this is not considered in the final image-based rendering evaluation and is not consistently used in research. This highlights the need for a robust evaluation framework which captures more than just the performance of the rendering network.

Subjective evaluation (using mean opinion scores) can also be performed to provide ground truth quality. However, in order to form robust performance benchmark statistics the use of human subjects can be prohibitive in terms of time. Most authors therefore only present selected rendered views, which are often limited in coverage. 

\paragraph{Scene Representation}\label{sec:related works}
INR research involves modelling different signal spaces as continuous representations. This encompasses paradigms such as modelling signal representations \cite{sitzmann2020implicit, saragadam2023wire} and interpolating visual space to improve resolution \cite{mundra2023livehand, chen2023cunerf}. Neural rendering falls within INR research as an automated 3-D graphical rendering approach. Currently, the most popular approach is NeRF rendering. To improve the visual quality from existing networks,  such as \cite{xu2022point} propose a ray marching and  approach to aggregate surface features for a neural 3-D point cloud. This is extended by Kulhanek et Sattler \cite{kulhanek2023tetra} who used tetrahedral representations. However, these methods modify neural representations rather than explicit representations.

In this paper, we aim to synthesise NeRF representations \textit{without a network} in order to avoid sampling bias in generating ground truth data. We leverage well-known transforms for ray-casting to generate radiance fields from mesh-based representations and modify subsequent representations using the same ray tracing functionality to provide realistic visual artifacts. Our method can be applied to other INRs, allowing us to evaluate various networks that could be involved in the NeRF pipeline.

\paragraph{Sampling Bias}\label{sec:depth}
Current evaluation methods lack a robust way of defining scene complexity relative to the distribution of training and the testing of ray samples. A bias is introduced when test samples originate from rays with position and direction parameters similar to samples drawn in training. Additionally it is not yet understood how the distribution of samples in 3-D space affects a network performance so this is rarely accounted for in classical NeRF evaluation.

To address this issue, we propose using a metric for scene complexity that takes account of the distribution of spatial parameters for each ray sample. We first select a set of training samples with a mean and standard deviation relative to  a normal distribution. We then introduce novel views as test data, which have different distributions to the training dataset. The change in standard deviations between the training and combined datasets is factored into our measure of task complexity. We assumed that the larger the standard deviation of novel views, the higher the probability that a network may fail to interpolate those views. Also, the range of the novel view distribution affects complexity. This is expressed algebraically in Section \ref{sec: scene complexity}.
Although a normal distribution can provide a reasonable approximation of scene complexity, it may not fully capture the skew that can occur in non-symmetric scenes. However, this issue is beyond the scope of this paper and remains a topic for future investigation.

\section{Proposed Methods for Evaluation}\label{sec:main}

\paragraph{NeRF Rendering} 
NeRF renders are accomplished by evaluating the colour and density of volumes sampled along a ray (\textit{ray samples}). Figure \ref{fig: Vol Rep} illustrates a single sample, where $t_i$ represents the depth of the sample relative to a view, located between the near and far collision planes, $t_n$ and $t_f$, respectively, and $\delta$ is the distance between $t_n$ and $t_f$. Each sample comprises the colour $c_i$ and the density $\sigma_i$ of a volume. $c_i$ and $\sigma_i$ can be written as $c_i(\mathbf{r_t}, \mathbf{d_k})$ and $\sigma_i(\mathbf{r_t})$, which are approximated using a neural network. The origin of the ray sample $\mathbf{r_t}$ is defined as $\mathbf{r_t} = \mathbf{o_k} + t_i \mathbf{d_k}$, where  $\mathbf{o_k}$ and $\mathbf{d_k}$ are the origin and direction vectors of a ray originating from a view. Following \cite{mildenhall2021nerf}, the colour $C_{k}$ of the pixel $k$ in a given view is rendered using Equation \ref{eq: NeRF colouring}.
\begin{equation}\label{eq: NeRF colouring}
    C_{k} = \sum_{i=1}^{N} T_i(1-\exp(-\sigma_{i} \delta_{i}))\mathbf{c_i}, \text{and} \: \: T_i = \exp (\sum_{j=1}^{i-1} \sigma_{j}\delta_k),
\end{equation}
\noindent where $T_i$ represents the transmittance of the $i$-th sample along each ray, $j$ represents a sample in-front of sample $i$, thus indicating the significance of each sample's visual parameters in the final render.

A network can be defined to estimate $c_i(\mathbf{r_t}, \mathbf{d_k})$ and $\sigma_i(\mathbf{r_t})$ with a continuous and differentiable function $F$ using Equation \ref{eq:inr} \cite{sitzmann2020implicit}, 
\begin{equation}\label{eq:inr}
    F(x,\phi, \nabla_x\phi, \nabla_x^2\phi, ...) = 0, \phi:x\rightarrow\phi(x).
\end{equation}
\noindent Here, $F$ is a function representing a scene and $\phi(x)$ is a network which estimates $F$. Mildenhall et al. \cite{mildenhall2021nerf} show that $\phi(x)$ can be modelled using a Multi-Layer Perceptron (MLP), where $t_i$ is uniformly sampled, thus $\delta_i$ is constant. It should be noted that $\phi \in R^n$ and Equation \ref{eq:inr} can be used to represent any INR; for the case of NeRF $\mathbf{x}$ is a 5-D input defined as $\mathbf{x} = (\mathbf{r_t}, \mathbf{d_k})$.

Due to imperfections of the sampling method and image pose estimation \cite{wang2023benchmarking, neff2021donerf}, any image quality metric used to evaluate the NeRF render inevitably accumulates  errors that arise across the entire NeRF pipeline, rather than errors that solely come from the rendering network. Additionally, since image-based metrics evaluate $C_{k}$ rather than $\sigma_i$ and $\delta_i$, information on prediction quality relative to the distribution of samples in space is lost.

\paragraph{Whole-scene Average Prediction Error (WAPE)} 
The proposed framework for evaluating INR network performance is shown in Figure \ref{fig:framework}. The process first generates ground truth volumetric representations. The modules used in this process can be adjusted for different representations. The information provided by selected methods for synthetic generation also enables us to evaluate the difficulty of learning different visual effects and address our concerns regarding sampling bias.

Because ground truth is available, we adopt mean absolute error (MAE) as the evaluation metric in  WAPE, as defined in Equation \ref{eq:WAPE}.
\begin{equation}\label{eq:WAPE}
    \epsilon(\xi) = MAE(\xi^* - \xi),
\end{equation}
\noindent where $\xi$ is a learnable parameter, and $\xi^*$ is the corresponding ground truth \footnote{Note that we tested both MAE and mean squared error (MSE) and found that MAE gave better results for both $\mathbf{c}$ and $\sigma$, given that $0 < \xi < 1$}.
For NeRFs, when the colour or density is evaluated, $\xi$ is $\mathbf{c}$ or $\sigma$, respectively. For other INRs, $\xi$ could be $t$, $t_n$ and $t_f$ or $\delta$. Multiple parameters can also be evaluated within the same experiment.

% \textcolor{red}{not sure I understand this sentence:}This enables concise evaluation with a variable set of parameters which can be evaluated simultaneously. 

%The ``whole-scene" should be considered based on the limitations of the scene suggested by the novel-view testing data. 
Because we can control the position of any view, the novel views can be placed around the region(s) in a scene where we expect a network to perform novel-view synthesis. For example, if we want to evaluate a NeRFs ability to infer higher resolution views we could place novel views close to objects within a scene.

We tested our method with two experiments: the first evaluates the performance of several well-known NeRF networks on a scene containing a range of visual complexities; the second demonstrates the configurability of our framework by evaluating different activation functions used in a coordinate MLP to learn the depth of a simple cuboid scene.

\paragraph{Task Complexity Metric}\label{sec: scene complexity}

Ideally, the complexity of a task should reflect the difficulty of learning a scene with respect to its visual features and the training parameters. However, no established procedure for evaluating such complexity currently exists for NeRFs. Thus here we have exploited the complexity of a ray tracer's integral component as an indicator of its ability to simulate reality. The integral component is expected to contain parameters which modify the complexity of the light fields by the chosen ray tracers. More precisely, the number of points along a trace and number of light samples\footnote{Which accounts for the number of traces casted along each ray and the quantisation factor for colour transformation, respectively.} are used
to derive $\rho \cdot ord$, where $\rho$ is the sample size multiplied by the number of light sources and $ord$ is the order of a trace's integral function. As a number of shaders, $N_{\text{shaders}}$, is generally compiled, we accumulated the ray tracing complexities using Equation \ref{eq: shader complexity}.
\begin{equation}\label{eq: shader complexity}
    \lambda = \sum_{w=0}^{N_{\text{shaders}}} \frac{256}{\Omega_w} (\rho_w \cdot ord_w),
\end{equation}
\noindent where $\Omega_w$ is a scaling factor which considers the influence of each shader relative to the maximum colour transformation (256 is used for 8-bit colour depth). Given that some shaders transform colours over a smaller range (e.g. using dimmer light-sources)  we assume that subtle variations introduced by these shaders are more difficult for a NeRF to capture.
As transformations are discretized to RGB values, some subtleties may be lost, so we only considered the maximum range of transformation in RGB form.

We also considered in our metric, the differences in the distributions of training and testing data; so an overall complexity $\Lambda$ was defined as in Equation \ref{eq: final complexity}.
\begin{equation}\label{eq: final complexity}
    \Lambda = n_{pts} \cdot \lambda \cdot |std_{\text{train}}-std_{\text{novel view}}|,
\end{equation}
\noindent where $n_{pts}$ is the number of training samples and $|std_{\text{train}}-std_{\text{novel view}}|$ is absolute difference of the standard deviation of ray-positions in a given training and testing data, respectively.

Whereas this definition of task complexity may be seen as simplistic \footnote{More investigation is required to better determine the properties of radiance-field representations that hinder learning.}, our framework can used to experiment on how the positional distribution of ray samples, not just the views, may affect task complexity. 
%It may be initially reasonable to assume that regions with fewer samples will be more difficult to interpolate. Such an exploration could help us develop more sophisticated metrics to evaluate the complexity of INR tasks.

%%%%%%%%%%%%%%%%%%%%%%%%%%%%%%%%%%%%%%%%%%%%%%%%%%%%%%%%%%%%%%%%%%%%%%%%%%%%%%%%%%%%%%%%%%%%%%%%%%%%%%%%%%%%%%%%%%%%%%%%%%%%%%%%%%%%%%%%%%%%%%%%%%%%%%%%%%%%%%%%%%%%%%%%%%%%%%%%%%%%%%%%%%%%%%%%%%%%%%%%%%%%%%%%%%%%%%%%%%%%%%%%%%%%%%%%%%

\section{Data Synthesis}\label{sec:sythesizing}
We generate ground truth data based on: (1) %outlining the algebra involved with 
transforming a triangular-surface mesh representations into explicit NeRF representations, and (2) modifying ray tracing algorithms to incorporate different light-field effects into the explicit representation.

\paragraph{From Meshes to NeRF Geometry}\label{sec: nerf geometry}
To transform a mesh-based geometry into NeRF geometry, we require the exact $\delta_i$ and $t_i$. We follow well-known procedures for ray casting. First, we redefine $t_i \in \Psi$ such that $t_{u,v}$ is an index in the matrix $\Psi$, where $u \in [0, s_s]$ identifies a triangular surface from a batch of surfaces with size $s_s$, $v \in [0,s_r]$ identifies a ray from a batch (e.g. a view) with size $s_r$ and $t_{u,v} = t_i$ represents a valid sample. Using parameters defining each ray ($\mathbf{o_v}$ and $\mathbf{d_v}$) and face-data (i.e. normals, $\mathbf{n_u}$, and positions of the plane along each normal, $k_u$), we apply the vector ray-plane intersection to find $t_{u,v}$ using Equation \ref{eq:t_o}
\begin{equation}\label{eq:t_o}
    t_{u,v} = \frac{- (\mathbf{o_v} \cdot \mathbf{n_u} +k_u)}{\mathbf{d_v} \cdot \mathbf{n_u}}
\end{equation}
We then mask $t_{u,v}$ where intersections fall within the surface bounds. For triangular meshes, the \textit{barycentric coordinate system} is used to find the constants $\alpha$, $\beta$ and $\gamma$ in $\mathbf{r_{u,v}} = \alpha\mathbf{a}+ \beta\mathbf{b} + \gamma\mathbf{c}$. Here $\mathbf{r_{u,v}} = \mathbf{r_t}$, $(\mathbf{a},\mathbf{b},\mathbf{c})$ are the vertices of the triangular surface and $\alpha > 0, \beta > 0$ and $\gamma >0$ are conditions for viable ray-plane intersection. $(\alpha, \beta, \gamma)$ are three unknowns which can be solved simultaneously using the equations pertaining the $(x, y, z)$ of each vertex. For rectangular meshes, linear boundary conditions can be placed on the coordinates of $r_{u,v}$.

Where $t_{u,v} \neq nan$ we define $\delta'_{u,v}$ using Equation \ref{eq:delta}. We choose a constant $\delta=0.001$ which was small enough to prevent $\delta'_{u,v} \rightarrow \infty$ at large angles between the ray and surface. In our experiments we found that this also means that the exponential term in Equation \ref{eq: NeRF colouring} will tend to be $1$. We consequently remove the $\delta_i$ from the exponential term but retain it in the transmittance term. This is because it is not required to link opacity to surface thickness, but it . is still important for defining transmittance.
\begin{equation}\label{eq:delta}
    \delta'_{u,v} =\delta\frac{|\mathbf{d_v}| |\mathbf{n_u}|}{\mathbf{d_v} \cdot \mathbf{n_u}}
\end{equation}
%We wrote the components $n_u, k_u, o_v, d_v$ into matrices and pre-computed the resulting matrices. This allowed us to access batch data for faster training, relative to accessible hardware\footnote{We used an NVIDIA RTX 3090 and were able to solve $1$ million intersections in less than a minute.}. Alternatively, there may be benefit in generating sample geometry during training. 

$\mathbf{c_{u,v}} \in \zeta$ and $\sigma_{u,v} \in \Theta$ can be used to represent the colour and density for the masked $\Psi$. By maintaining a record of the valid indices we can obtain the spatial locations, $r_{u,v}$, of valid intersections and their relative colour and density; which are also reused in our approach to shading. It should be noted that we initialized $c_{u,v} = c_{0}$ and $\sigma_{u,v} = \sigma_{0}$ where $c_0$ and $\sigma_0$ are the initial ray sample colours without light-features. 

\paragraph{Shading NeRF Geometry}\label{sec:vis artifacts}
We apply the classical rendering equation using ray tracing as described in \cite{immel1986radiosity, kajiya1986proceedings} to provide two material shaders for rendering explicit NeRFs. Ray tracing estimates the trajectory of a ray, emitted from a given light-source and reflected $ord_w$ times before being reaching a view. By accumulating the colour of the surface and sampling the light intensity at each point of reflection, we can simulate a variety of realistic visual features. For example, diffuse material effects only consider the first point of reflection, while reflective materials will consider a longer trace. The shaders can be applied directly to $\zeta$ and reflected traces are found by solving the intersections of reflected rays using Equation \ref{eq:t_o} and the barycentric coordinate conditions, where $\mathbf{o_v}' = \mathbf{r_{u,v}}$ and $\mathbf{d_v}' = \mathbf{d_v} - 2 \frac{\mathbf{d_v} \cdot \mathbf{n_u}}{\mathbf{n_u} \cdot \mathbf{n_u}} \mathbf{n_u}$.

Shaders utilise the principles of \textit{solid angles} to sample light intensity from a singular source. Our first shader uses the lambertian diffuse \textit{bidirectional reflectance distribution function} (BRDF) equation \cite{nicodemus1965directional} to accumulate colour and light intensity. It is configured to provide different light intensities for solid materials defined where $\sigma_{u,v} = 1$ and glass-like materials where $\sigma_{u,v} < 1$. The second shader models reflections for the glass material. %We have decided not to pursue this further as it deviates from the scope of our paper.

%%%%%%%%%%%%%%%%%%%%%%%%%%%%%%%%%%%%%%%%%%%%%%%%%%%%%%%%%%%%%%%%%%%%%%%%%%%%%%%%%%%%%%%%%%%%%%%%%%%%%%%%%%%%%%%%%%%%%%%%%%%%%%%%%%%%%%%%%%%%%%%%%%%%%%%%%%%%%%%%%%%%%%%%%%%%%

\section{Experiments and Discussion}\label{sec: experiments}
In this section, we use the metric for task complexity $\Lambda$ to meet specific experimental conditions. In the first experiment, it enables us to compare the performance of networks learning two sets of visual features. In the second experiment, the complexity metric was used to ensure that we evaluated INRs with optimal performance. Specifically, we set $\Lambda = 0$ to indicate the easiest task possible for a given scene. We found this metric to be especially suitable for NeRF as subtleties in visual complexity were amplified by the $n_{pts}$ parameter.

\subsection{Novel View Synthesis}\label{sec:large model}
In this experiment, we evaluated the performance of different neural rendering networks, i.e. INGP, Mip-NeRF \cite{barron2021mip}, and Nerfacto, using the scene shown in Figure \ref{fig:framework} (top-left).  INGP and Mip-NeRF provide fast and high-quality results, respectively. Nerfacto is part of the NeRFStudio pipeline which we have adapted for evaluation. %We used all the shaders previously presented and varied the use of the reflection shader.

\begin{figure}[t]
    \centering
    \includegraphics[width=\textwidth]{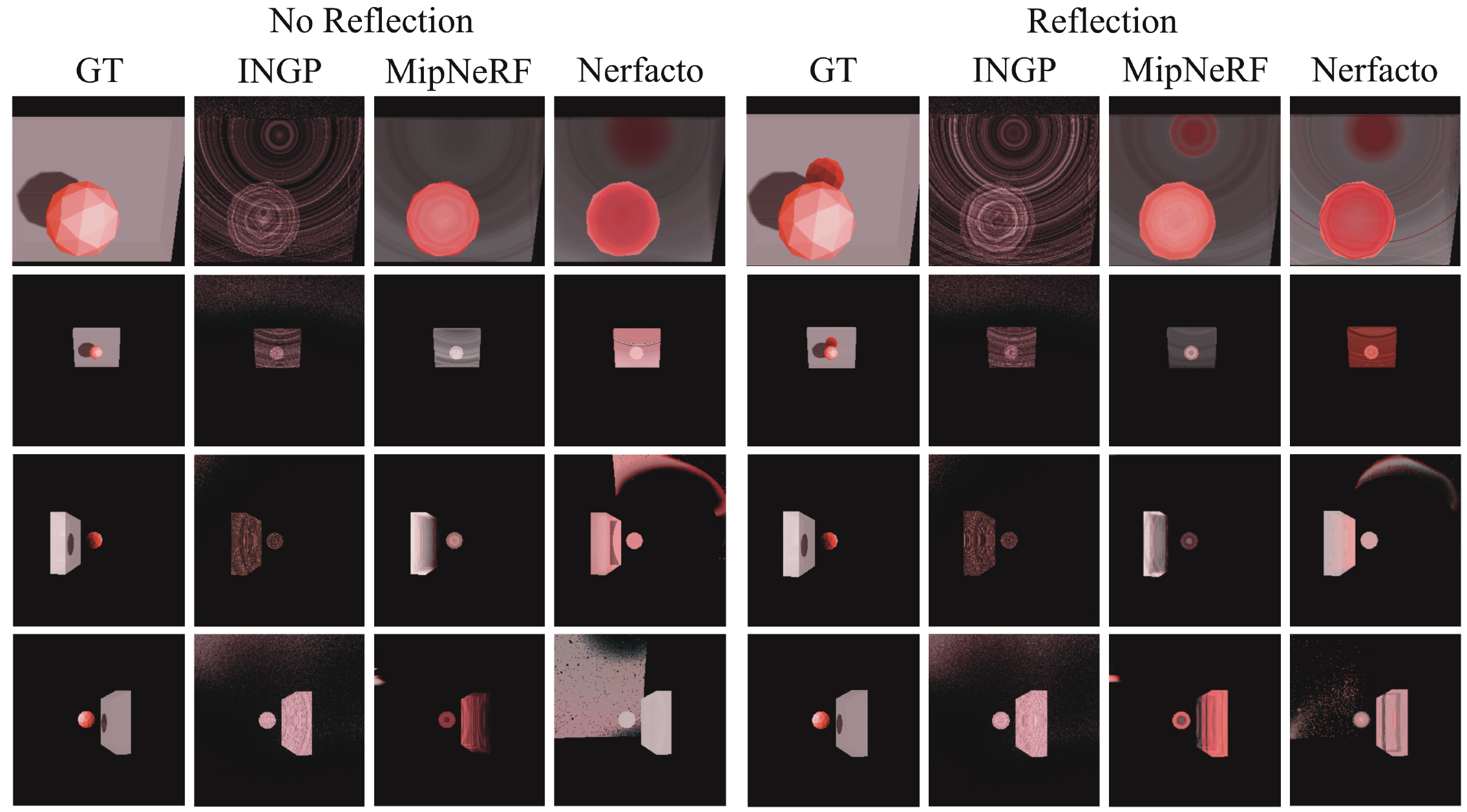}
    \caption{Novel view rendering results without (left) and with (right) the reflection shader.}
    \label{fig: nerf result}
\end{figure}

The test scene employed contains twenty views with $250\times250$ resolution and four novel views positioned at different distances from the ball object. The important parameters and complexity measurements are shown in Table \ref{tab:params and metrics}. Figure \ref{fig: nerf result} illustrates the rendering results of the novel views shown as blue cameras in Figure \ref{fig:framework} (top-left). Performance comparisons shown in Table \ref{tab:nerf results} reveal that Mip-NeRF outperforms other methods, indicated by WAPE, SSIM and LPIPS.
The subjective results in Figure \ref{fig: nerf result} confirm this as Mip-NeRF method can differentiate the colours of separate objects without corrupting the predictions of empty space.  Figure \ref{fig: nerf result} also shows that INGP and Nerfacto are not suitable for sparse datasets. Additionally, the INGP model tended to over-train and failed to capture color (more subjective results can be found in Supplementary Materials \ref{sec: additional experiments}). The WAPE results indicate that Nerfacto performs better than INGP in predicting colour, but both struggle with density predictions.

\begin{table}[t]
\caption{Parameters used for data synthesising and training. More parameters in Supplementary Materials \ref{sec: training params}.}
    \centering
    \begin{tabular}{c cc cc cc}
    \toprule
        &Paramaters & Values & Paramaters & Values & Metrics & Values \\\hline
        \multirow{4}{*}{Synthesizing} & $\rho_{\text{diffuse}}$,  $\rho_{\text{glass}}$ & 25 & $\rho_{\text{reflection}}$ & 1 & $\lambda_{\text{no reflection}}$ & 50.0 \\
        &$ord_{\text{diffuse}}$,  $ord_{\text{glass}}$ & 1 & $ord_{\text{reflection}}$ & 2 & $\lambda_{\text{reflection}}$ & 54.0 \\
        &$\Omega_{\text{diffuse}}$,  $\Omega_{\text{glass}}$ & 255 & $\Omega_{\text{reflection}}$ & 2 & $\Lambda_{\text{no reflection}}$ & $3.68\times10^{8}$ \\
        &$n_{pts}$ & $1.04\times10^{6}$ & & & $\Lambda_{\text{reflection}}$ & $3.98\times10^{8}$ \\ \hline
        Training &$\text{loss}_{\text{rgb}}$ & MSE & $\text{loss}_{\text{density}}$ & L1 & &  \\
    \bottomrule
    \end{tabular}
    
    \label{tab:params and metrics}
\end{table}
\begin{table}[t]
\caption{Results from using WAPE for colour and density predictions and PSNR, SSIM, LPIPS, where R indicates the use of the reflection shader.}
    \centering
    \begin{tabular}{l ccccc c}
    \toprule
        Model & \multicolumn{2}{c}{WAPE} & PSNR & SSIM & LPIPS & R \\  \cmidrule(r){2-3}
        & $c$ & $\sigma$ & & & \\\hline
        INGP     & $0.250 \pm 0.001$ & $0.183 \pm 0.003$ & $14.9 \pm 0.1$ & $0.52 \pm 0.01$ & $0.41 \pm 0.01$ & No \\
        Mip-NeRF & \cellcolor{green!25} $0.081 \pm 0.003$ & \cellcolor{green!25} $0.049 \pm 0.009$ & $17.9 \pm 0.4$ & \cellcolor{green!25}$0.88 \pm 0.01$ & \cellcolor{green!25} $0.18 \pm 0.01$ & No \\
        Nerfacto & $0.113 \pm 0.011$ & $0.134 \pm 0.002$ & \cellcolor{green!25}$18.4 \pm 1.4$ & $0.82 \pm 0.02$ & $0.26 \pm 0.01$ & No \\ \hline
        INGP     & $0.242 \pm 0.004$ & $0.181 \pm 0.005$ & $14.8 \pm 0.1$ & $0.69 \pm 0.00$ & $0.40 \pm 0.00$ & Yes\\
        Mip-NeRF & \cellcolor{green!25}$0.068 \pm 0.004$ & \cellcolor{green!25}$0.022 \pm 0.004$ & \cellcolor{green!25} $24.2 \pm 0.6$ & \cellcolor{green!25} $0.87 \pm 0.01$ & \cellcolor{green!25} $0.16 \pm 0.01$ & Yes\\
        Nerfacto & $0.111 \pm 0.021$ & $0.148 \pm 0.005$ & $17.4 \pm 1.1$ & $0.81 \pm 0.03$ & $0.24 \pm 0.01$ & Yes \\ 
        
    \bottomrule
    \end{tabular}
    \label{tab:nerf results}
\end{table}

Output parameters are conventionally trained by rendering predictions with Equation \ref{eq: NeRF colouring} with an MSE loss function. Because of the existence of ground truth data in our method, we were able to train the predictions explicitly. We observed that $c$ and $\sigma$ learnt better with separate loss functions, as seen from training results in Supplementary Materials \ref{sec: additional experiments}. Furthermore we found the predictions for $c$ converge much faster than $\sigma$. Even for INGP, the $\sigma$ predictionsconverged $~1000$ training iterations later which resulted in over-training $c$ before $\sigma$ could converge.

We observe that the WAPE metric provides more information than the other metrics. For example, in Table \ref{tab:nerf results}, although PSNR results indicate that the Nerfacto method is superior, the uncertainty of PSNR values ($\pm$5.7) suggests that Nerfacto may have trouble with prediction consistency. On the other hand, the high uncertainty of WAPE colour value implies that the Nerfacto method struggles with approximating colour consistently despite having an lower score for its colour predictions than its density predictions when compared to the INGP and Mip-NeRF WAPE results.

\subsection{Evaluating depth estimated by INRs}\label{sec:eval depth inr}

This experiment investigated the adaptability of our framework for evaluating other INRs under different conditions. We generated an explicit representation of the spatial parameter $t_{u,v}$ relative to a scene containing three \textit{rectangular} cuboid meshes and three neighbouring views. To adapt the mesh-transformation to rectangular surfaces we used rectangular boundary conditions rather than the barycentric coordinate conditions. We sampled training and novel view data from the same three views (with a data split of $0.8$, respectively) to evaluate the case where the whole-scene is limited to the region of the training views. This bias reduced the complexity of the scene to approximately $0$ which allows us to evaluate networks with their optimal capability. This could be helpful to indicate which networks should be best suited for the task of view synthesis. Additionally, we allowed the network to learn only the earliest occurring sample along each ray to ensure that geometric complexities remain close to $0$.

\begin{figure}[t]
    \centering
    \includegraphics[width=\textwidth]{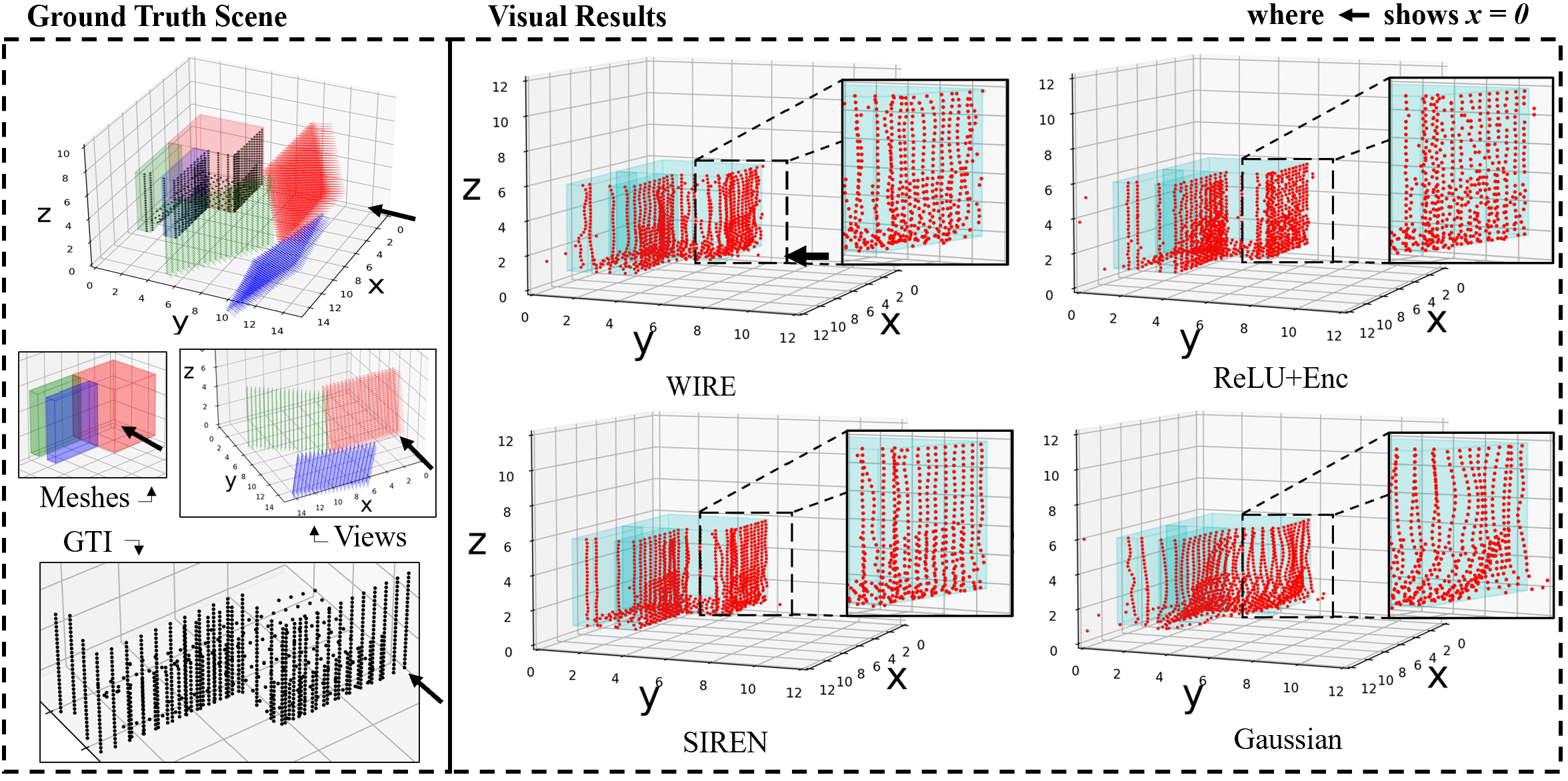}
    \caption{\textbf{Ground Truth Scene:} We show three meshes (blue, green, red) which are intersected by rays from three views (shown as blue, green and red batches of arrows), where the ground truth intersections (GTI) illustrates the known point-cloud locations of ray-surface intersections. \textbf{Visual Results:} We visualise the predictions of GTI (red points) relative to all the surfaces in our scene (in blue).}
    \label{fig:depth vis results}
\end{figure}

We compared WAPE with PSNR of the 3-D spatial predictions by evaluating four activation functions: (1) ReLU  and spatial encoders used in \cite{mildenhall2021nerf}, (2) \textit{Wavelet Implicit Neural Representations} (WIRE), (3) \textit{Sinusoidal Representation Networks} (SIREN) and (4) \textit{Gaussian Activated Radiance Fields} (GARF) \cite{saragadam2023wire,sitzmann2020implicit,chng2022garf}, using the same network architectures as presented in \cite{saragadam2023wire}. The mesh representation and subsequent surface predictions are visualised in Figure \ref{fig:depth vis results}, where we used $200\times200$ pixel view size. The result comparison is shown in Table \ref{tab:inr results}.

\begin{table}
\caption{Performance of depth by INRs ($\xi = t_{u,v}$) evaluated with different objective quality metrics. Green and red indicate the best and the worst performances, respectively.}
    \centering
    \begin{tabular}{ccccc}
    \toprule
         & ReLU & GAUSS & WIRE & SIREN \\\hline
         PSNR & $44.0 \pm 9.6$ &  $42.3 \pm 6.0$ & \cellcolor{green!25}$44.5 \pm 10.6$ & \cellcolor{red!25} $42.0 \pm 2.1$ \\
         WAPE & $0.101 \pm 0.024$ & \cellcolor{red!25}$0.185 \pm 0.019$ & $0.109 \pm 0.021$ & \cellcolor{green!25} $0.092 \pm 0.017$ \\
         \bottomrule
    \end{tabular}
    
    \label{tab:inr results}
\end{table}

As shown in Figure \ref{fig:depth vis results} (right), SIREN produced the smoothest estimate of the panels (low-frequency features). This finding is consistent with the results in Table \ref{tab:inr results} which demonstrate the superior performance of SIREN based on the WAPE metric. The Gaussian method also produced a relatively smooth result, but heavily underestimated a significant portion of the scene, which could be the reason why it achieved a higher PSNR than SIREN and lower WAPE than SIREN. The ReLU+Enc method generated noisy results and struggles with fine precision placement. This result coheres with limitations of the ReLU frequency encoding presented in \cite{mildenhall2021nerf}. Out of all results, SIREN visually gave the best structured result (high-frequency areas), which aligns with the WAPE metric.

%%%%%%%%%%%%%%%%%%%%%%%%%%%%%%%%%%%%%%%%%%%%%%%%%%%%%%%%%%%%%%%%%%%%%%%%%%%%%%%%%%%%%%%%%%%%%%%%%%%%

\section{Conclusions, Limitations and Future Work}\label{sec: limits}

This paper presents a new approach that demonstrates the potential of using ray-based algorithms to perform objective parametric evaluation. This supports evaluation of INRs, as well as  provides insight into challenges faced during network training. 
Moreover, our conceptualisation of task complexity highlights factors that can influence learning which are not available from current methods. For example, we can now objectively compare the difficulties of learning different sets of visual and geometric features.

We believe that this new framework provides a strong platform for further investigation. In particular, the paradigm for optimal NeRF image capture. It is understood that visually appealing NeRFs require a lot of images with a sufficient spatial distribution; it would thus be valuable to design a strategy that optimises view selection while minimising the number of necessary images, and considers physical obstructions such that one could theoretically replicate the shots in real life. Notably, this could inform us on the optimal placement of novel views for evaluation. This would enable necessary amendments to our task complexity metric, ensuring the training difficulty is better represented.

One limitation of our method is the high computational cost due to large matrices such as $\Psi, \zeta, \Theta$, particularly for high-resolution complex scenes.
As our data synthesis employs only triangular or rectangular meshes, we cannot use classical datasets which include a large variety and sum of polygons. Those datasets are however designed for generating image data and not explicit representations.
This highlights the necessity of creating more advanced benchmarks, which take into account the combination of geometric and visual features present in typical real-world scenes. In order to create a useable WAPE metric, future research should focus on developing scalable benchmarks, allowing for objective comparisons across different training views and resolutions. 
%While we did not include large-scale scenes in our experiments due to memory and computational cost required for processing, 
This scalability can achieved by exploiting ray-based view synthesis tools, such as Unity.

%% Resting Place %%
% [  ]

% %%%%%%%%%%%%%%%%%%%%%%%%%%%%%%%%%%%%%%%%%%%%%%%%%%%%%%%%%%%%
\bibliographystyle{unsrt}
\bibliography{neurips_2023}

\begin{thebibliography}{10}

\bibitem{mildenhall2021nerf}
Ben Mildenhall, Pratul~P Srinivasan, Matthew Tancik, Jonathan~T Barron, Ravi
  Ramamoorthi, and Ren Ng.
\newblock \text{NeRF}: Representing scenes as neural radiance fields for view
  synthesis.
\newblock {\em Communications of the ACM}, 65(1):99--106, 2021.

\bibitem{gao2022nerf}
Kyle Gao, Yin Gao, Hongjie He, Denning Lu, Linlin Xu, and Jonathan Li.
\newblock \text{NeRF}: Neural radiance field in 3d vision, a comprehensive
  review.
\newblock {\em ArXiv}, abs/2210.00379, 2022.

\bibitem{xie2022neural}
Yiheng Xie, Towaki Takikawa, Shunsuke Saito, Or~Litany, Shiqin Yan, Numair
  Khan, Federico Tombari, James Tompkin, Vincent Sitzmann, and Srinath Sridhar.
\newblock Neural fields in visual computing and beyond.
\newblock In {\em Computer Graphics Forum}, volume 41(2), pages 641--676. Wiley
  Online Library, 2022.

\bibitem{tancik2023nerfstudio}
Matthew Tancik, Ethan Weber, Evonne Ng, Ruilong Li, Brent Yi, Justin Kerr,
  Terrance Wang, Alexander Kristoffersen, Jake Austin, Kamyar Salahi, Abhik
  Ahuja, David McAllister, and Angjoo Kanazawa.
\newblock Nerfstudio: A modular framework for neural radiance field
  development.
\newblock {\em ArXiv}, abs/2302.04264, 2023.

\bibitem{levy2023seathru}
Deborah Levy, Amit Peleg, Naama Pearl, Dan Rosenbaum, Derya Akkaynak, Simon
  Korman, and Tali Treibitz.
\newblock \text{SeaThru-NeRF}: Neural radiance fields in scattering media.
\newblock {\em ArXiv}, abs/2304.07743, 2023.

\bibitem{jambon2023nerfshop}
Cl{\'e}ment Jambon, Bernhard Kerbl, Georgios Kopanas, Stavros Diolatzis, George
  Drettakis, and Thomas Leimk{\"u}hler.
\newblock \text{NeRFshop}: Interactive editing of neural radiance fields.
\newblock {\em Proceedings of the ACM on Computer Graphics and Interactive
  Techniques}, 6(1), 2023.

\bibitem{kong2023vmap}
Xin Kong, Shikun Liu, Marwan Taher, and Andrew~J Davison.
\newblock \text{vMAP}: Vectorised object mapping for neural field slam.
\newblock {\em ArXiv}, abs/2302.01838, 2023.

\bibitem{chng2022garf}
Shin-Fang Chng, Sameera Ramasinghe, Jamie Sherrah, and Simon Lucey.
\newblock \text{GARF}: Gaussian activated radiance fields for high fidelity
  reconstruction and pose estimation.
\newblock {\em ArXiv}, abs/2212.02280, 2022.

\bibitem{yu2021plenoxels}
Alex Yu, Sara Fridovich-Keil, Matthew Tancik, Qinhong Chen, Benjamin Recht, and
  Angjoo Kanazawa.
\newblock Plenoxels: Radiance fields without neural networks.
\newblock {\em ArXiv}, abs/2112.05131, 2021.

\bibitem{kurz2022adanerf}
Andreas Kurz, Thomas Neff, Zhaoyang Lv, Michael Zollh{\"o}fer, and Markus
  Steinberger.
\newblock \text{AdaNeRF}: Adaptive sampling for real-time rendering of neural
  radiance fields.
\newblock In {\em Computer Vision--ECCV 2022: 17th European Conference, Tel
  Aviv, Israel, October 23--27, 2022, Proceedings, Part XVII}, pages 254--270.
  Springer, 2022.

\bibitem{neff2021donerf}
Thomas Neff, Pascal Stadlbauer, Mathias Parger, Andreas Kurz, Joerg~H Mueller,
  Chakravarty R~Alla Chaitanya, Anton Kaplanyan, and Markus Steinberger.
\newblock Donerf: Towards real-time rendering of compact neural radiance fields
  using depth oracle networks.
\newblock In {\em Computer Graphics Forum}, volume 40(4), pages 45--59. Wiley
  Online Library, 2021.

\bibitem{wang2023benchmarking}
Chen Wang, Angtian Wang, Junbo Li, Alan Yuille, and Cihang Xie.
\newblock Benchmarking robustness in neural radiance fields.
\newblock {\em ArXiv}, abs/2301.04075, 2023.

\bibitem{muller2022instant}
Thomas M{\"u}ller, Alex Evans, Christoph Schied, and Alexander Keller.
\newblock Instant neural graphics primitives with a multiresolution hash
  encoding.
\newblock {\em ACM Transactions on Graphics (ToG)}, 41(4):1--15, 2022.

\bibitem{yu2021plenoctrees}
Alex Yu, Ruilong Li, Matthew Tancik, Hao Li, Ren Ng, and Angjoo Kanazawa.
\newblock Plenoctrees for real-time rendering of neural radiance fields.
\newblock In {\em Proceedings of the IEEE/CVF International Conference on
  Computer Vision}, pages 5752--5761, 2021.

\bibitem{mildenhall2022nerf}
Ben Mildenhall, Peter Hedman, Ricardo Martin-Brualla, Pratul~P Srinivasan, and
  Jonathan~T Barron.
\newblock Nerf in the dark: High dynamic range view synthesis from noisy raw
  images.
\newblock In {\em Proceedings of the IEEE/CVF Conference on Computer Vision and
  Pattern Recognition}, pages 16190--16199, 2022.

\bibitem{zhang2020nerf++}
Kai Zhang, Gernot Riegler, Noah Snavely, and Vladlen Koltun.
\newblock \text{NeRF}++: Analyzing and improving neural radiance fields.
\newblock {\em ArXiv}, abs/2010.07492, 2020.

\bibitem{wang2004image}
Zhou Wang, Alan~C Bovik, Hamid~R Sheikh, and Eero~P Simoncelli.
\newblock Image quality assessment: from error visibility to structural
  similarity.
\newblock {\em IEEE transactions on image processing}, 13(4):600--612, 2004.

\bibitem{zhang2018unreasonable}
Richard Zhang, Phillip Isola, Alexei~A Efros, Eli Shechtman, and Oliver Wang.
\newblock The unreasonable effectiveness of deep features as a perceptual
  metric.
\newblock In {\em Proceedings of the IEEE conference on computer vision and
  pattern recognition}, pages 586--595, 2018.

\bibitem{de2023scannerf}
Luca De~Luigi, Damiano Bolognini, Federico Domeniconi, Daniele De~Gregorio,
  Matteo Poggi, and Luigi Di~Stefano.
\newblock \text{ScanNeRF}: a scalable benchmark for neural radiance fields.
\newblock In {\em Proceedings of the IEEE/CVF Winter Conference on Applications
  of Computer Vision}, pages 816--825, 2023.

\bibitem{lee2023dense}
Dongwoo Lee and Kyoung~Mu Lee.
\newblock Dense depth-guided generalizable nerf.
\newblock {\em IEEE Signal Processing Letters}, 30:75--79, 2023.

\bibitem{uy2023scade}
Mikaela~Angelina Uy, Ricardo Martin-Brualla, Leonidas Guibas, and Ke~Li.
\newblock Scade: Nerfs from space carving with ambiguity-aware depth estimates.
\newblock {\em ArXiv}, abs/2303.13582, 2023.

\bibitem{sitzmann2020implicit}
Vincent Sitzmann, Julien Martel, Alexander Bergman, David Lindell, and Gordon
  Wetzstein.
\newblock Implicit neural representations with periodic activation functions.
\newblock {\em Advances in Neural Information Processing Systems},
  33:7462--7473, 2020.

\bibitem{saragadam2023wire}
Vishwanath Saragadam, Daniel LeJeune, Jasper Tan, Guha Balakrishnan, Ashok
  Veeraraghavan, and Richard~G Baraniuk.
\newblock Wire: Wavelet implicit neural representations.
\newblock {\em ArXiv}, abs/2301.05187, 2023.

\bibitem{mundra2023livehand}
Akshay Mundra, Jiayi Wang, Marc Habermann, Christian Theobalt, Mohamed
  Elgharib, et~al.
\newblock \text{LiveHand}: Real-time and photorealistic neural hand rendering.
\newblock {\em ArXiv}, abs/2302.07672, 2023.

\bibitem{chen2023cunerf}
Zixuan Chen, Jianhuang Lai, Lingxiao Yang, and Xiaohua Xie.
\newblock {CuNeRF: Cube-Based Neural Radiance Field for Zero-Shot Medical Image
  Arbitrary-Scale Super Resolution}.
\newblock {\em ArXiv}, abs/2303.16242, 2023.

\bibitem{xu2022point}
Qiangeng Xu, Zexiang Xu, Julien Philip, Sai Bi, Zhixin Shu, Kalyan Sunkavalli,
  and Ulrich Neumann.
\newblock \text{Point-NeRF}: Point-based neural radiance fields.
\newblock In {\em Proceedings of the IEEE/CVF Conference on Computer Vision and
  Pattern Recognition}, pages 5438--5448, 2022.

\bibitem{kulhanek2023tetra}
Jonas Kulhanek and Torsten Sattler.
\newblock \text{Tetra-NeRF}: Representing neural radiance fields using
  tetrahedra.
\newblock {\em ArXiv}, abs/2304.09987, 2023.

\bibitem{immel1986radiosity}
David~S Immel, Michael~F Cohen, and Donald~P Greenberg.
\newblock A radiosity method for non-diffuse environments.
\newblock {\em Acm Siggraph Computer Graphics}, 20(4):133--142, 1986.

\bibitem{kajiya1986proceedings}
James~T Kajiya.
\newblock Proceedings of the 13th annual conference on computer graphics and
  interactive techniques, 1986.

\bibitem{nicodemus1965directional}
Fred~E Nicodemus.
\newblock Directional reflectance and emissivity of an opaque surface.
\newblock {\em Applied optics}, 4(7):767--775, 1965.

\bibitem{barron2021mip}
Jonathan~T Barron, Ben Mildenhall, Matthew Tancik, Peter Hedman, Ricardo
  Martin-Brualla, and Pratul~P Srinivasan.
\newblock \text{Mip-NeRF}: A multiscale representation for anti-aliasing neural
  radiance fields.
\newblock In {\em Proceedings of the IEEE/CVF International Conference on
  Computer Vision}, pages 5855--5864, 2021.

\end{thebibliography}

\appendix

\newpage

\begin{center}

\textbf{\Large Towards a Robust Framework for NeRF Evaluation}

\textbf{\large Supplementary Material}
\end{center}

\section{Implementation Details}

\subsection{Configuring Sample Representations}

\paragraph{Section 5.1 Novel View Synthesis}
We set $t_{u, v} \leq 1000$.
Ground truth ray samples were defined where all valid surface intersections lie along a ray. Additionally, in cases where a ray contained no valid surface intersections a ray sample was defined at the end of the ray with $\sigma_{u,v} = 0$. This denoted the representation of empty space at the intersection with the scene's bounding box.

\paragraph{Section 5.2 Evaluating depth estimated by INRs}
We set $t_{u, v} \leq 100$.
Ground truth ray samples were defined by the earliest occurring intersection. Similarly, rays without intersections contained a sample at the furthest position along the bounded ray.

\subsection{Training Environments}
\paragraph{Section 5.1 Novel View Synthesis}
We employed the Nerfstudio training environment due to its user-friendly interface and effective management. This provides a number of functions for modelling cameras and rays and handling rendering. This is discussed in our code.

\paragraph{Section 5.2 Evaluating depth estimated by INRs}
We used the neural network implementations provided by \cite{saragadam2023wire} and trained the MLPs in a simple testing script.

\subsection{Training Parameters} \label{sec: training params}
\paragraph{Section 5.1 Novel View Synthesis}
We used the default hyper-parameters of each network, as provided by Nerfstudio. We used $5,000$,  training iterations for INGP, $12,000$ for Nerfacto and $25,000$ for Mip-NeRF. We then selected the best training results using the PSNR, SSIM and LPIPS metrics to inform our selection. This ensured that the comparison to the WAPE results was not biased by what the WAPE metric indicated was the point of optimal performance. Interestingly, we found that training INGP with a batch size of $256$ ray samples yielded  optimal results, while Nerfacto and Mip-NeRF both required a batch size of $4096$. This was done to prevent INGP from over training which happened much faster with larger batches. For similar reason, INGP used a learning rate of $1e-4$, while Nerfacto and Mip-NeRF used a learning rate of $1e-2$ and $5e-4$, respectively. Note that in both cases the hyperparameters for Nerfacto and Mip-NeRF are similar to those selected in their respective papers. 

\paragraph{Section 5.2 Evaluating depth estimated by INRs}
We tuned the hyper-parameters of each MLP to ensure that the condition for optimal performance was met. The default MLP has $5$ layers with $256$ nodes. For the WIRE activated MLP this was reduced by $\frac{\text{\# of nodes}}{\sqrt{2}} = 181$ nodes as discussed in \cite{saragadam2023wire}. These were trained for $1500$ epochs with a batch size of $256$ and learning rate of $1e-5$ using the \textit{Adam} optimiser and MSE loss. For the ReLU activation function, $4$ frequencies were used for the positional encoding. For the SIREN activation, the optimal signal scale was $2.5$. For the Gaussian activation, the optimal signal bandwidth was $1.2$. For WIRE the optimal signal scale and bandwidth was $1.4$ and $1.0$, respectively.

\section{Additional Experiments and Discussion} \label{sec: additional experiments}
\subsection{Novel-view Synthesis}
\begin{figure}[t]
     \centering
     \begin{subfigure}[b]{\textwidth}
         \centering
         \includegraphics[width=0.8\textwidth]{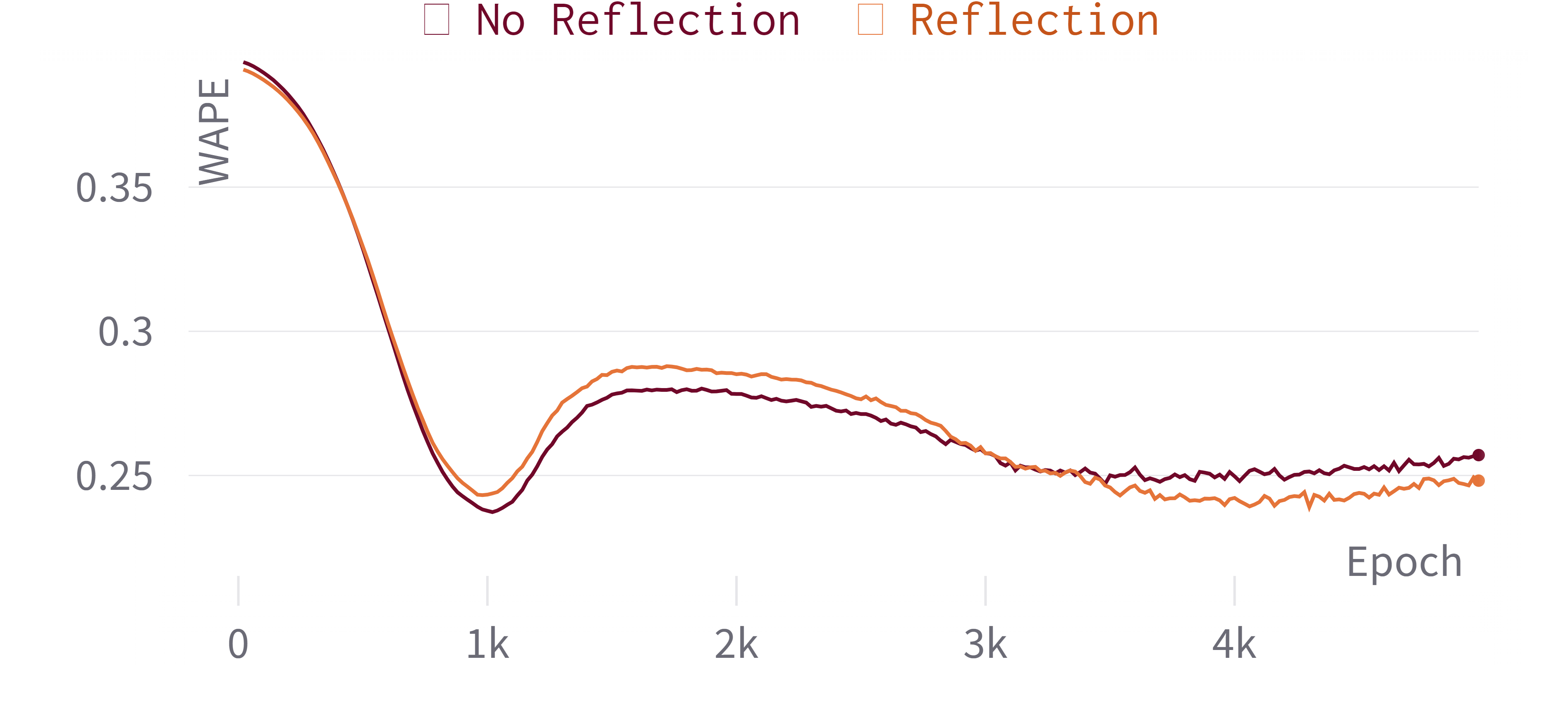}
         \caption{WAPE for $\xi=c$}   \vspace{3mm}
         \label{fig:wapecol-ingp}
     \end{subfigure}
     % \hfill
     \begin{subfigure}[b]{\textwidth}
         \centering
         \includegraphics[width=0.8\textwidth]{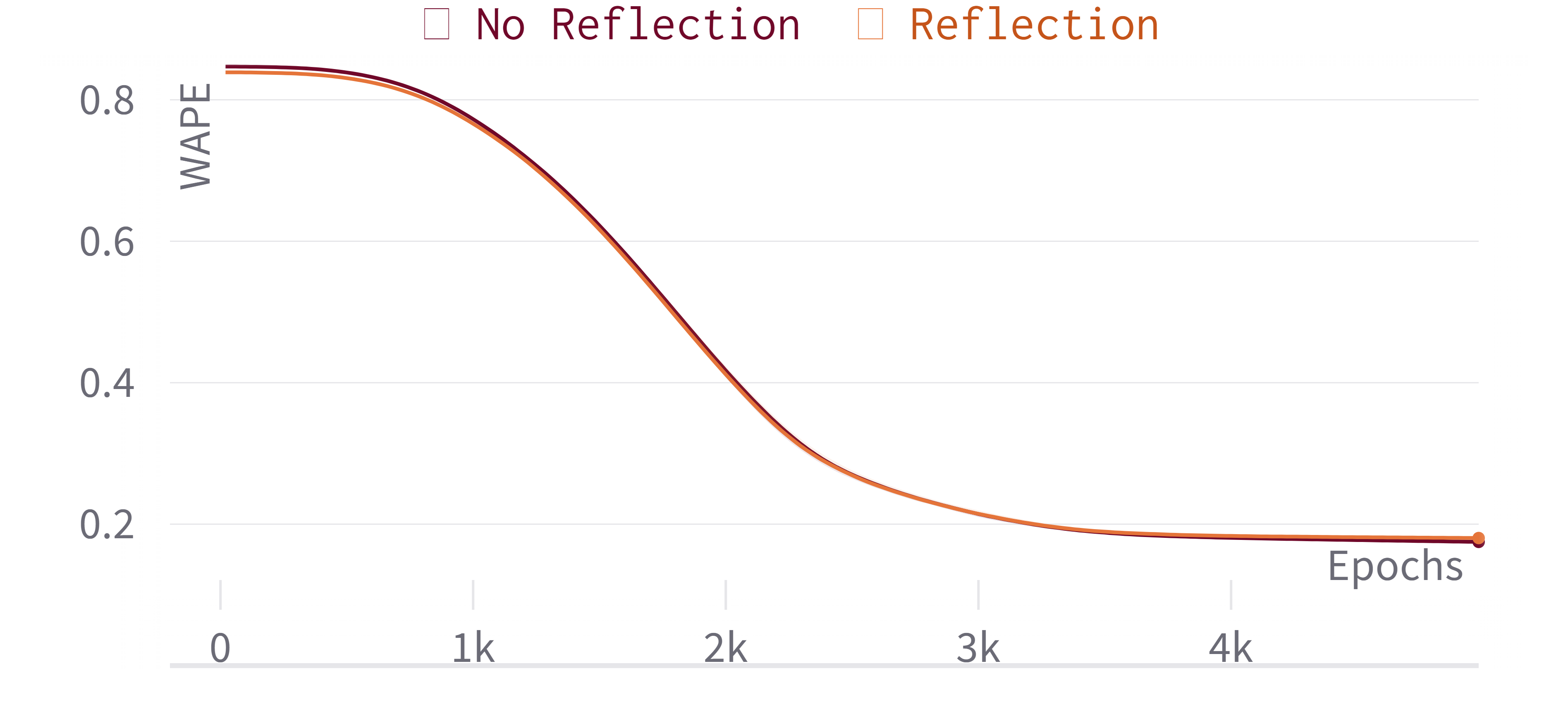}
         \caption{WAPE for $\xi=\sigma$}
         \label{fig:wapesig-ingp}
     \end{subfigure}
        \caption{WAPE results during the training process of the INGP network, where the \textit{Reflection} (orange) and \textit{No Reflection} (black) plots demonstrate the results of using the reflection shader.}
        \label{fig:ingp training graphs}
\end{figure}

\begin{figure}[t]
    \centering
    \includegraphics[width=\textwidth]{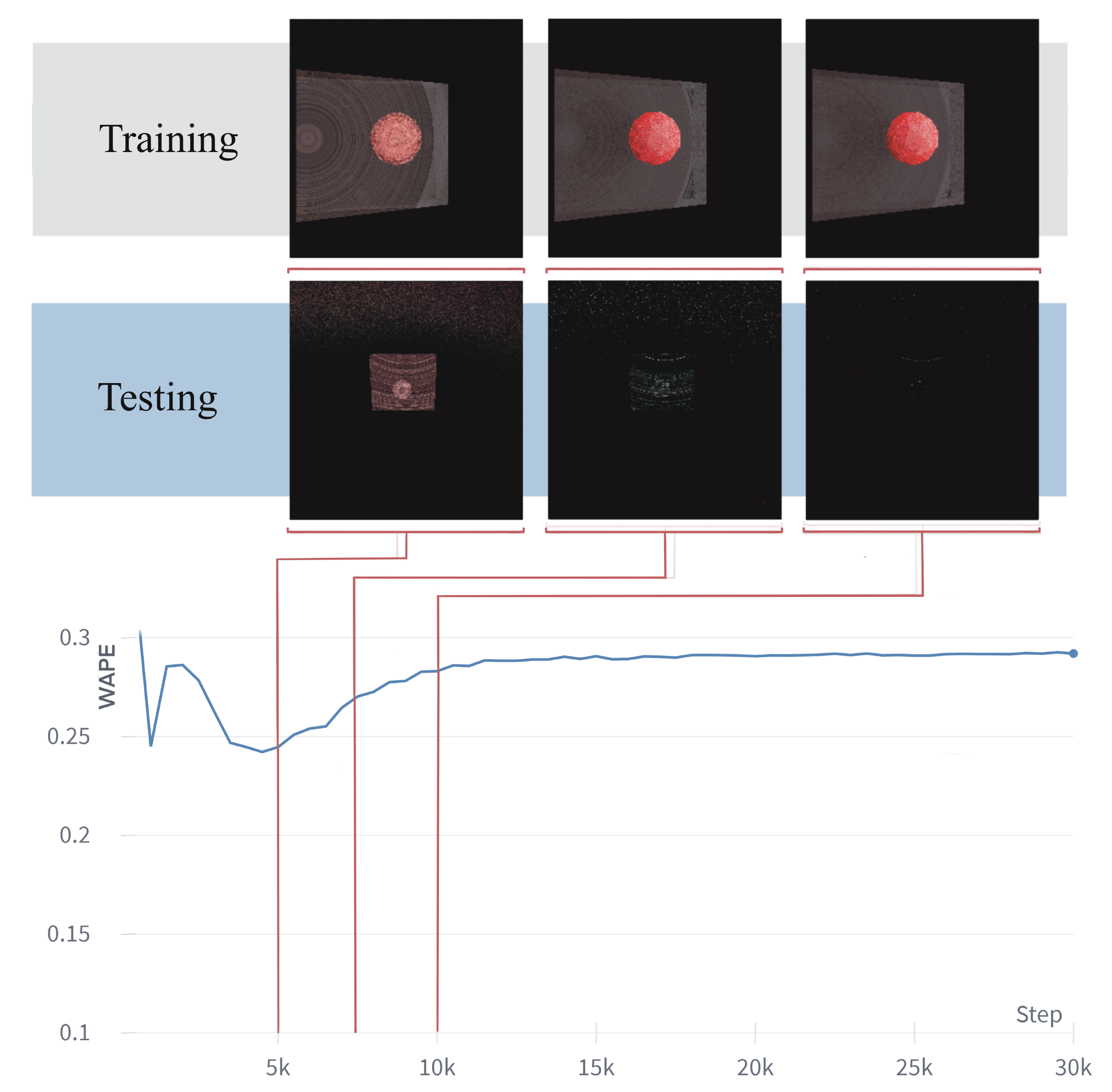}
    \caption{\textbf{Visual Results:} Training (grey) and novel view (blue) testing results taken at epochs $5,000$, $7,500$ and $10,000$. \textbf{WAPE Plot:} WAPE $\xi=c$ results (blue) for INGP over $30,000$ epochs.}
    \label{fig: overtraing visualisation}
\end{figure}

\paragraph{INGP Overtraining} \label{sec: suppmat ingp overtrain}
Figure \ref{fig:ingp training graphs} shows the WAPE metric for the $c$ and $\sigma$ predictions during the training. At epoch $1000$, the error of the colour prediction reaches a minimum, while the density prediction begins to converge at a higher rate. During the subsequent period of training, the colour predictions begin to diverge, while the density predictions continue to decrease. Around epoch $4000$, the colour prediction error reaches another minimum and then begin to diverge again. This behaviour indicates that training colour and density end-to-end negatively affects performance. This leads to overtraining as shown in Figure \ref{fig: overtraing visualisation}, where shortly after epoch $4000$, the parametric colour prediction converges to a larger error value as training view quality increases. 

\begin{figure}[t]
     \centering
     \begin{subfigure}[b]{\textwidth}
         \centering
         \includegraphics[width=0.75\textwidth]{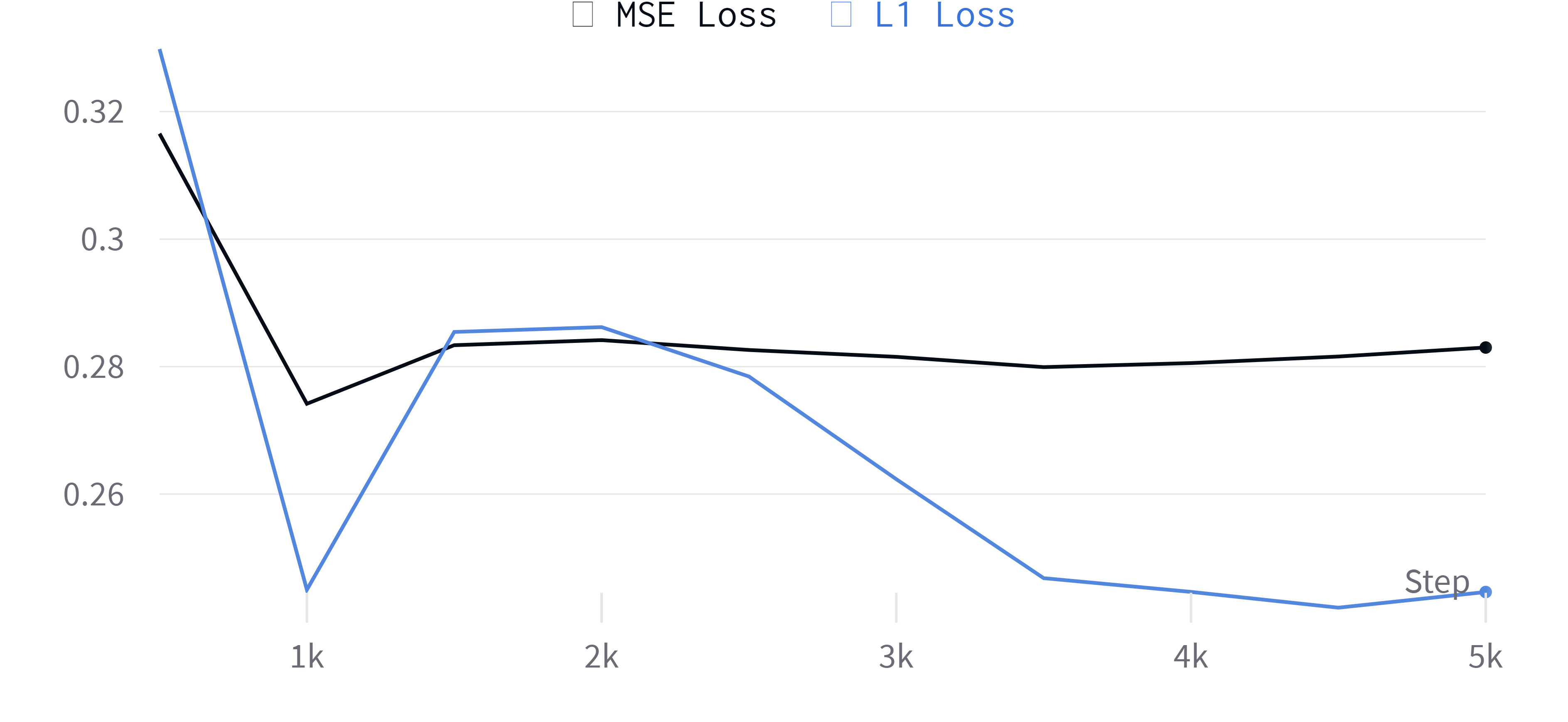}
         \caption{WAPE for $\xi=c$} \vspace{3mm}
         \label{fig:wapecol-ingp 2}
     \end{subfigure}
     % \hfill
     \begin{subfigure}[b]{\textwidth}
         \centering
         \includegraphics[width=0.75\textwidth]{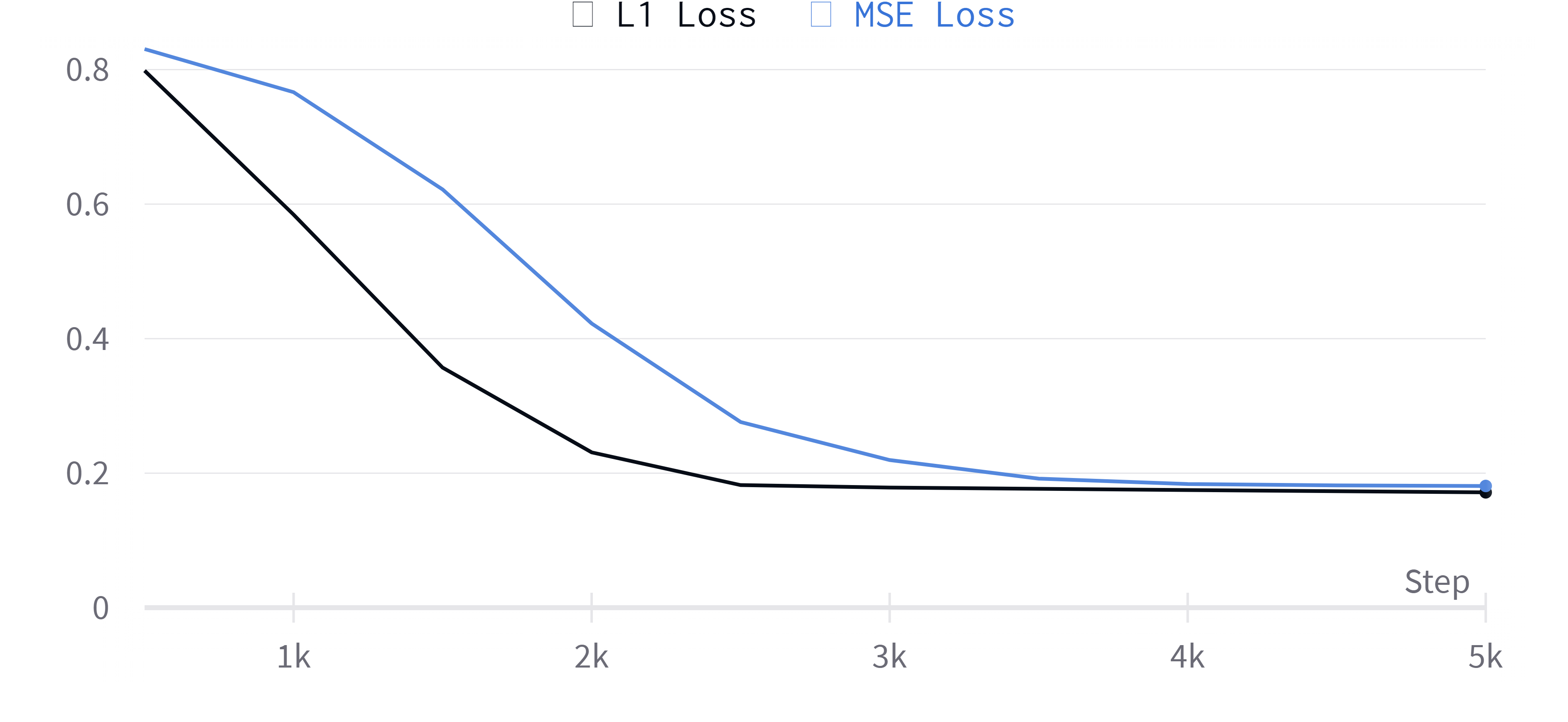}
         \caption{WAPE for $\xi=\sigma$}
         \label{fig:wapesig-ingp 2}
     \end{subfigure}
        \caption{WAPE results during the training process of the INGP network with different loss functions}
        \label{fig:ingp loss graphs}
\end{figure}

\paragraph{Parametric Loss Functions}
Our WAPE scores show that the MSE loss function was best suited for optimising the colour predictions, while the L1 loss function was found to be more effective for optimising the density predictions. Traditionally, NeRFs employ an MSE loss function to train pixel-colour predictions relative to ground truth image values. Therefore, it is not surprising that the MSE loss function works well for training the parametric colour predictions. Furthermore, using the L1 loss function to train the colour predictions results in no learning. Differently, Figure \ref{fig:ingp loss graphs} shows that the L1 loss function was the most effective for training the density output parameter as it has significantly less impact on training the colour parameters. This behaviour was also found with the Nerfacto and Mip-NeRF method, which may indicate a flaw with training the colour and density predictions end-to-end.

Interestingly, using the MSE loss function to train the density predictions suppresses the problems with overtraining INGP. However, this results in significantly worse colour predictions as it converges much earlier.

\subsection{Depth estimated INRs}
\begin{figure}
    \centering
    \includegraphics[width=\textwidth]{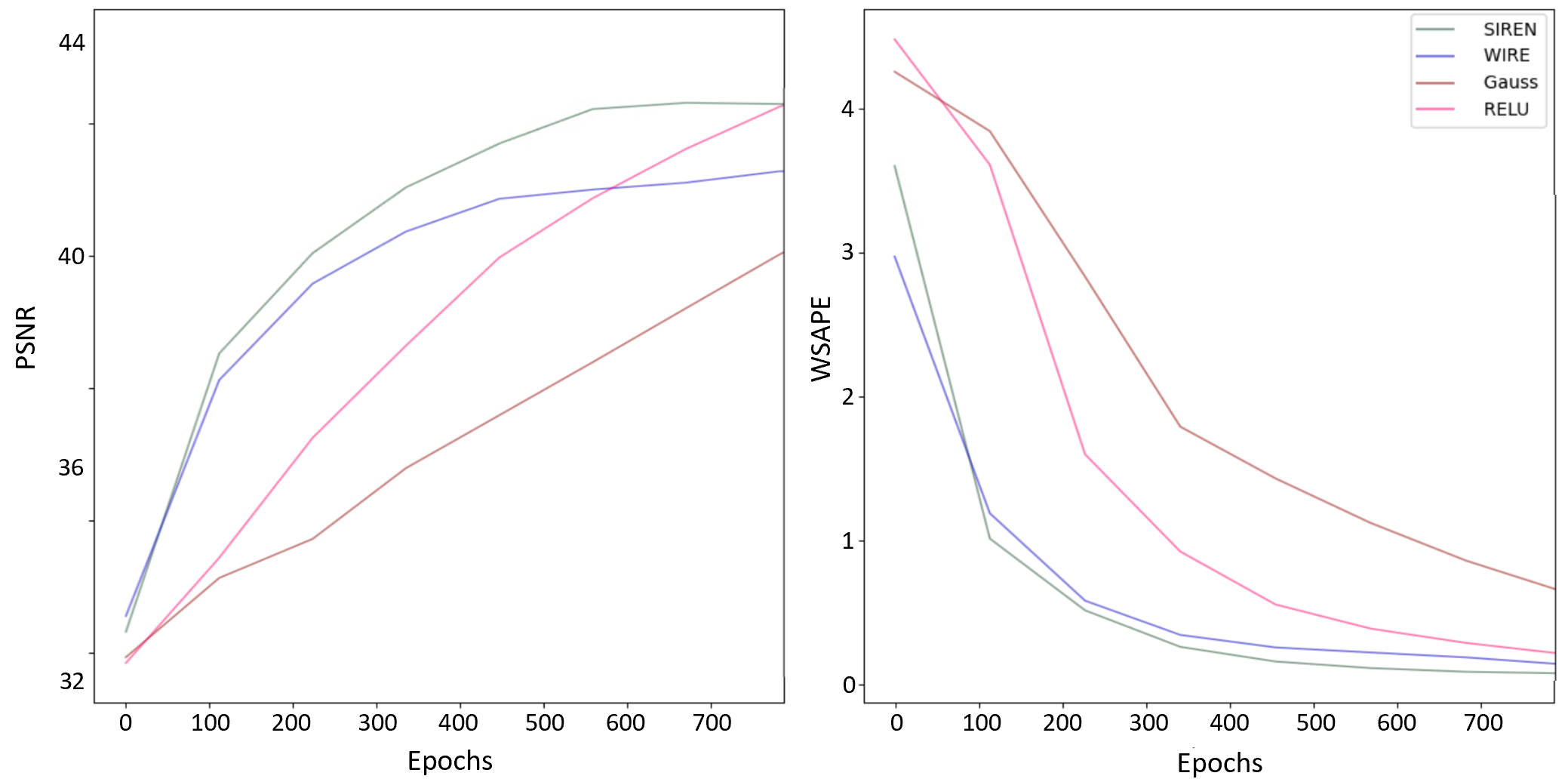}
    \caption{WAPE $\xi=t$ and 3-D PSNR results during training for depth estimation by INRs using different activation functions, incuding The ReLU (pink), Gaussian (brown), WIRE (blue) and SIREN (green).}
    \label{fig:depthinr training results}
\end{figure}

\paragraph{PSNR vs WAPE}
Figure \ref{fig:depthinr training results} presents the results of each metric during the first half of training. The Gaussian and ReLU activated networks converge significantly slower, indicating further potential of using either SIREN or WIRE. 

% Additionally, the results indicate a lack of correlation between the PSNR of 3-D spatial predictions and the WAPE. 
The PSNR metric is useful for assessing the absence of noise in a representation. However, this can be problematic when high frequency features dominate a scene, as shown in Figure \ref{fig:depthinr training results}. This demonstrates the benefit of using WAPE over PSNR for parametric evaluation.

\subsection{Simplified Hot-dog Dataset}

This experiment aims to highlight the advantage of using the WAPE method to investigate the effects of interpolating proximal novel views relative to a target object (in this case the hot-dog). This is demonstrated by reviewing the effects of the density and colour predictions on the final result. Consequently, the experiments provide more informative discussion on the challenges of modelling with NeRFs. 

We modified the well-known hot-dog dataset \cite{mildenhall2021nerf} by reducing the number of objects and surface sub-divisions in Blender (i.e. less planes), as shown in Figure \ref{tab:hotdog nerf results}. We used the same shaders as the reflection test in Section \ref{sec:large model}, thus the scene has a shading complexity of $\lambda = 54.0$. As $n_{pts} = 3.47 \times 10^{5}$ the task complexity is $\Lambda=1.88 \times 10^7$.

\begin{figure}
    \centering
    \includegraphics[width=\textwidth]{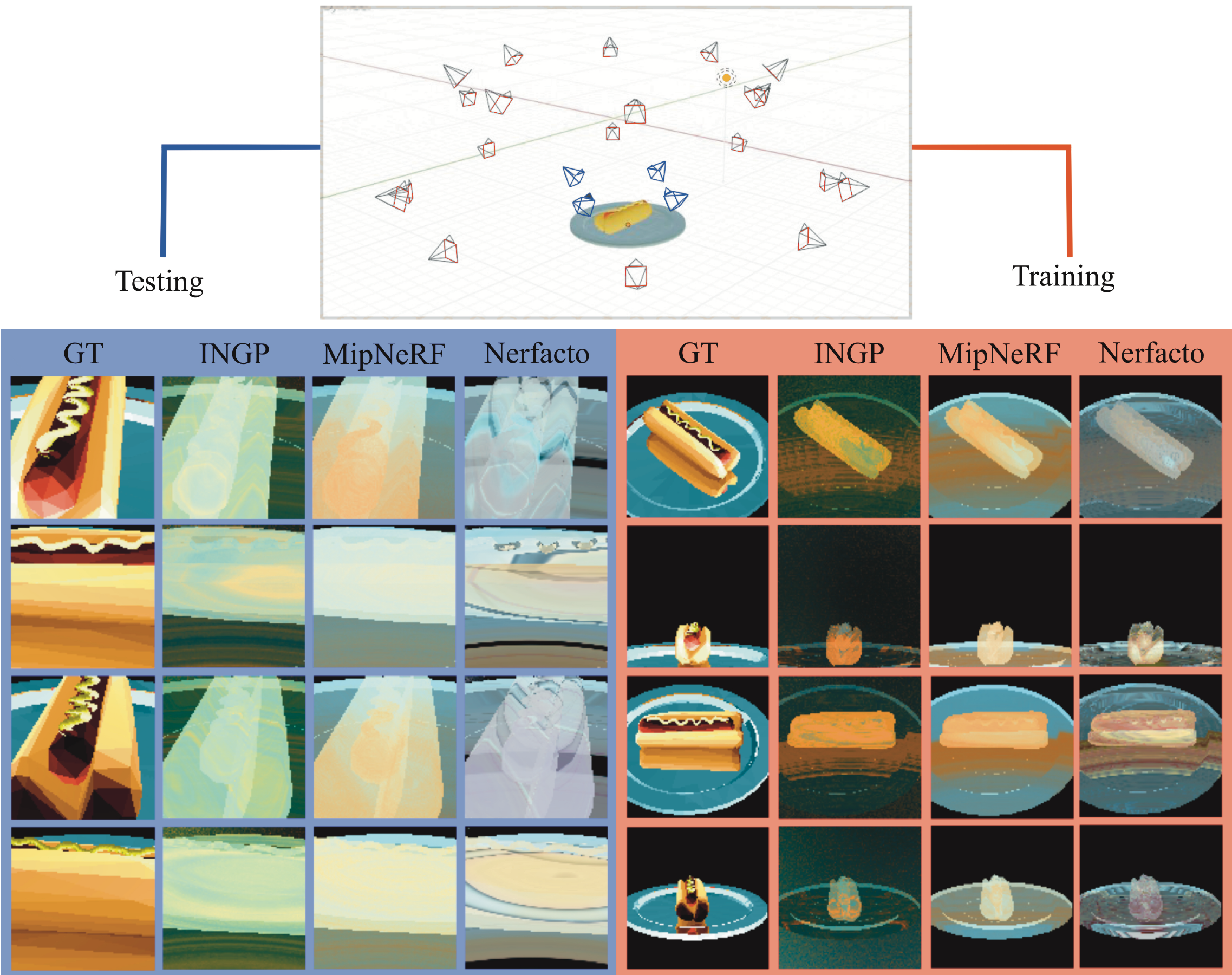}
    \caption{\textbf{Scene:} The Blender model where the twenty red cameras indicate the position of the training view, the four blue cameras indicate the position of the testing views. The position of the light-source is shown by the yellow point. \textbf{Visual Results:} The blue box presents the ground truth images and predicted novel views from testing each network. The red box presents four randomly selected training views for comparison.}
    \label{fig: hotdog visual results}
\end{figure}

The predictions of each novel-view are shown in Figure \ref{fig: hotdog visual results} and the numeric results are presented in Table \ref{tab:hotdog nerf results}. Comparing the training and testing views, we find that none of these models are capable of reaching a visually accurate prediction. This scene is subjectively more complex than the experiment in Section \ref{sec:large model}, though as the viewing resolution is smaller the task complexity of the hotdog scene is also less. This accounts for higher complexity in more detailed (higher resolution) scenes. However, the results indicate the need for additional investigations into the types of geometric complexities which make modelling a scene more challenging. Note that as both experiments use around $20$ training views, there is already cause to explore the effects of sparse-view synthesis in such investigations. The approximate sparsity of views that a network may be able to handle has yet to be studied.

Overall, the Mip-NeRF achieves the best results, which agrees with the results in Section \ref{sec:large model}. Differently, the WAPE $\sigma$ results and relative error margin indicate that the Nerfacto is significantly worse at predicting density for this experiment. This could be a consequence of using more materials, which would indicate that the Nerfacto method is much less suited for general use. %Though, more scenes and shaders need to be tested before this can be proven.

\begin{table}[t]
\caption{Results from the hot-dog test.} 
    \centering
    \begin{tabular}{l ccccc}
    \toprule
        Model & \multicolumn{2}{c}{WAPE} & PSNR & SSIM & LPIPS \\  \cmidrule(r){2-3}
        & $c$ & $\sigma$ & & & \\\hline
        INGP     & \cellcolor{red!25}$0.299 \pm 0.002$ & $0.105 \pm 0.001$ & \cellcolor{red!25} $3.7 \pm 0.0$ & \cellcolor{red!25} $0.30 \pm 0.00$ & \cellcolor{red!25} $0.76 \pm 0.01$ \\
        Mip-NeRF & \cellcolor{green!25} $0.252 \pm 0.006$ & \cellcolor{green!25} $0.092 \pm 0.005$ & \cellcolor{green!25} $10.0 \pm 0.2$ & \cellcolor{green!25} $0.37 \pm 0.001$ & \cellcolor{green!25} $0.65 \pm 0.02$ \\ 
        Nerfacto & $0.290 \pm 0.016$ & \cellcolor{red!25} $0.128 \pm 0.033$ & $9.6 \pm 0.3$ & $0.31 \pm 0.04$ &  $0.65 \pm 0.01$ \\
        
    \bottomrule
    \end{tabular}
    \label{tab:hotdog nerf results}
\end{table}

\subsection{Cube Cluster Dataset}
\begin{figure}
    \centering
    \includegraphics[width=\textwidth]{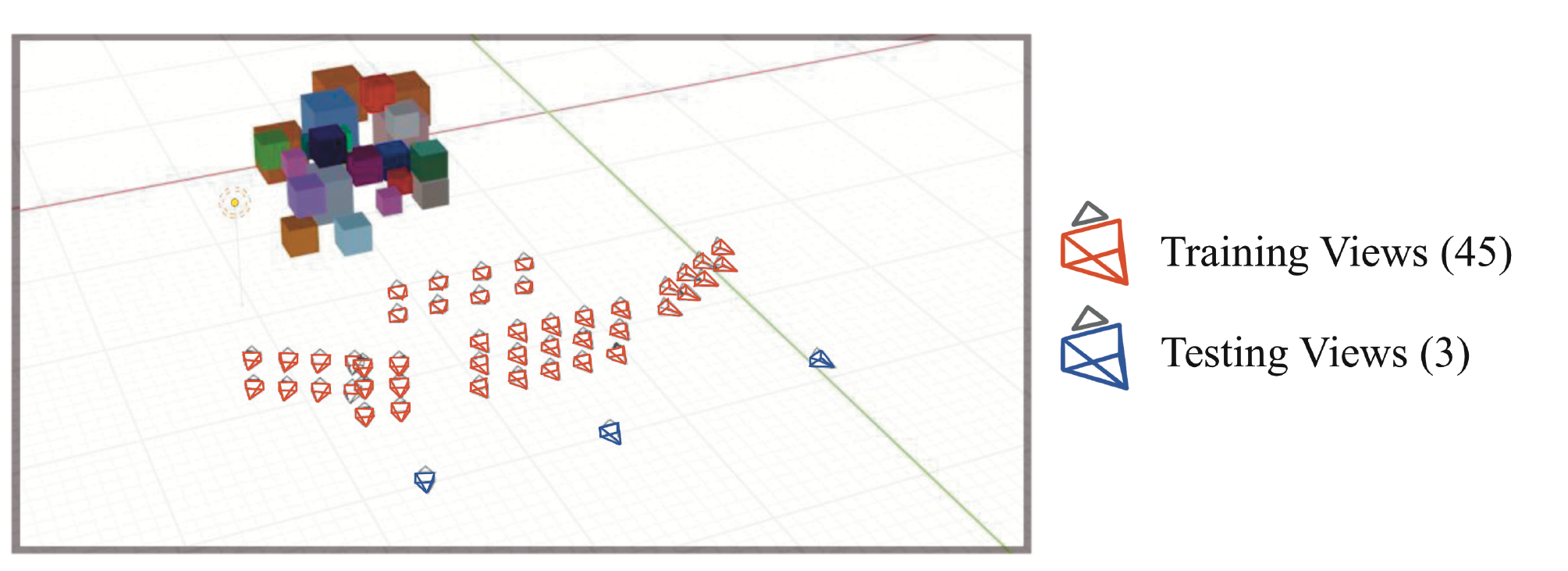}
    \caption{\textbf{Objects:} A cluster of cubes containing a set of reflective and non-reflective materials. \textbf{Cameras:} Red cameras indicated the training views and blue cameras indicate the testing views.\textbf{Light source:} A point light was placed to the left of the scene relative to the aggregate direction of the views.}
    \label{fig:cube vis results}
\end{figure}

\begin{figure}
    \centering
    \includegraphics[width=\textwidth]{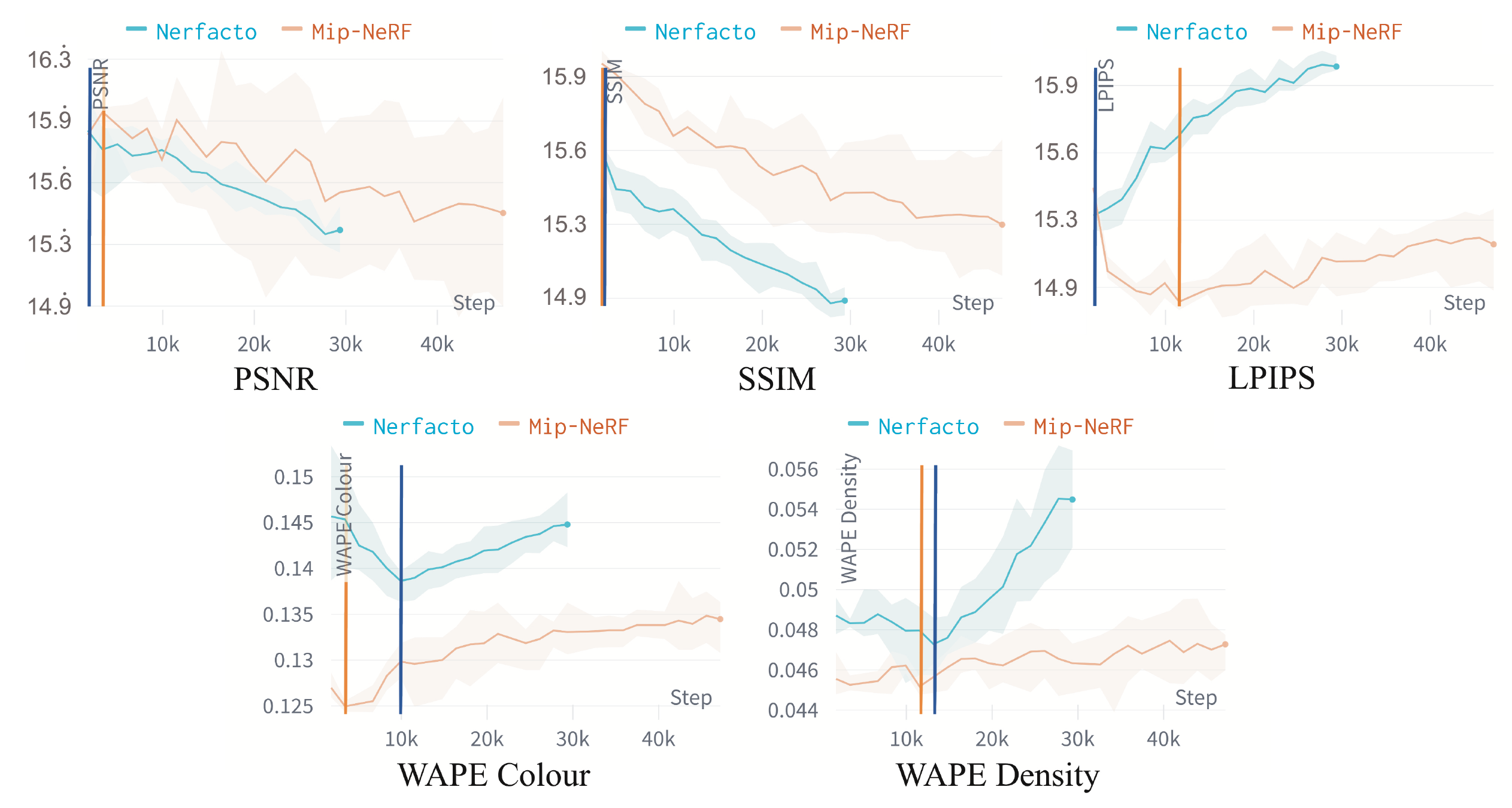}
    \caption{PSNR, SSIM, LPIPS and WAPE colour ($c$) and density ($\sigma$) plots during training where the blue vertical bar indicates the optimum epoch for the Nerfacto method and the orange vertical bar indicates the optimal epoch for the Mip-NeRF Method. }
    \label{fig:cube plot results}
\end{figure}

In this experiment we separately evaluated the use of PSNR, SSIM, LPIPS and WAPE for selecting the epoch where optimal performance is reached during training. To accomplish this we evaluated the Mip-NeRF and Nerfacto networks on the scene shown in Figure \ref{fig:cube vis results}. The scene contained $45$ training views and $3$ novel views with a resolution of $200 \times 200$ pixels and $\Lambda = 2.7 \times 10^7$. Each network was evaluated every 1000 epochs. The results are shown in Figure \ref{fig:cube plot results}, where the vertical lines highlight the optimal result as indicated by the relative metric.

Figure \ref{fig:cube plot results} illustrates the challenge of using the conventional metrics for image quality assessment. Almost all PSNR, SSIM and LPIPS results indicate that the optimal performance is achieved at the beginning of training, which is not expected. Conversely, the WAPE metrics for density and colour indicate that optimal performance happens $12000$ epochs later. The outliers to this are the Mip-NeRF results for LPIPS and WAPE $c$. In the first case, LPIPS indicates that epoch $12000$ is best. This agrees with the other WAPE results shown in Figure \ref{fig:cube plot results}. In the second case, the WAPE $c$ plot indicates that epoch $4000$ is best, which agrees with the PSNR result for Mip-NeRF. Despite this, the WAPE method computes errors for both colour and density predictions simultaneously so potentially incorrect results can be avoided by considering both error values. Moreover, PSNR and SSIM have large error margins. For PSNR the uncertainty is significantly large for Mip-NeRF that it envelopes the Nerfacto result.

\end{document}